\documentclass[lettersize,journal]{IEEEtran}
\usepackage{amsmath,amssymb,amsfonts}
\usepackage{algorithmic}
\usepackage{graphicx}
\usepackage{textcomp}
\usepackage{url}
\usepackage{array}
\usepackage{booktabs}
\usepackage[ruled,norelsize,vlined,linesnumbered]{algorithm2e}
\makeatletter
\newcommand{\removelatexerror}{\let\@latex@error\@gobble}
\makeatother
\usepackage{amssymb}
\usepackage{subfig}



\newtheorem{remark}{Remark}

\begin{document}

\title{Offline Reinforcement Learning for Wind Farm Control: A Wind Tunnel Study under Dynamic Wind Directions}

\author{Yuhan~Su,
        Hongyang~Dong,
        Simone~Tamaro,
        Filippo~Campagnolo,
        Carlo~L.~Bottasso,
        and~Xiaowei~Zhao
\thanks{This work was supported in part by the Horizon Europe Research and Innovation Actions under Grant Agreement 101122329 and in part by the UKRI Horizon Europe Guarantee Scheme under Grant 10095874. \textit{(Corresponding author: Hongyang Dong.)}}
\thanks{Yuhan Su, Hongyang Dong, and Xiaowei Zhao are with the Intelligent Control and Smart Energy (ICSE) Research Group, School of Engineering, University of Warwick, Coventry CV4 7AL, U.K. (e-mail: Yuhan.Su.1@warwick.ac.uk; hongyang.dong@warwick.ac.uk; xiaowei.zhao@warwick.ac.uk).}
\thanks{Simone Tamaro, Filippo Campagnolo, and Carlo L. Bottasso are with the Wind Energy Institute, Technical University of Munich, 85748 Garching bei M\"{u}nchen, Germany (e-mail: simone.tamaro@tum.de; filippo.campagnolo@tum.de; carlo.bottasso@tum.de).}
\thanks{This work has been submitted to the IEEE for possible publication. Copyright may be transferred without notice, after which this version may no longer be accessible.}
}

\markboth{Journal of \LaTeX\ Class Files,~Vol.~14, No.~8, August~2021}%
{Su \MakeLowercase{\textit{et al.}}: Offline Reinforcement Learning for Wind Farm Control}


\maketitle
\begin{abstract}
This paper addresses the wind farm power maximization problem in the presence of wind direction changes. Specifically, a model-free Modified Twin Delayed Deep Deterministic Policy Gradient with Behavior Cloning (MTD3-BC) algorithm is proposed to tackle this task through yaw control under varying wind direction conditions. MTD3-BC is an offline reinforcement learning (RL) algorithm that aims to infer good behavior from only a precollected offline dataset. Additionally, to ensure smooth and moderate yaw adjustments, a new action consistency term is introduced into the policy optimization objective. Unlike online RL methods, MTD3-BC does not require extensive interactions with a wind farm simulator during training, significantly reducing computational costs and training time. A wind tunnel experiment is conducted to validate the effectiveness of the algorithm under varying wind directions. The results demonstrate that MTD3-BC successfully mitigates wake effects, delivering farm-level power gains of approximately 10\% over the baseline greedy strategy and performance on par with a data-calibrated model-based wake-steering benchmark, while requiring no wake model and only a small fraction of the training cost of online RL. To our knowledge, this is the first time an offline RL wind farm control policy has been validated and demonstrated experimentally.
\end{abstract}


%
\IEEEpeerreviewmaketitle


\section{Introduction}
\label{se1}

\IEEEPARstart{W}{ind} energy has gained global attention for its crucial role in generating green and renewable energy, contributing to the reduction of carbon emissions and the pursuit of sustainable development. Several countries are increasing their investment in wind energy research and are expanding their wind energy capacity to gradually replace energy from fossil fuels. However, challenges remain in designing effective control laws due to the inherent complexity of wind energy systems. One major issue is the aerodynamic interactions among turbines, which give rise to  wake effects \cite{gonzalez2012wake,shakoor2016wake}. In fact, the wakes generated by upstream turbines, due to their reduced wind speed, lead to decreased power generation in downstream turbines \cite{howland2022collective}. It is worth noting that the operational optimization of a wind farm is, in practice, a broader problem than power capture alone, encompassing structural loads, component lifetime, curtailment requirements, and other grid-integration constraints. Power boosting through coordinated control is thus one particular mode of operation. The present work addresses this specific mode, targeting farm-level power capture through wake steering under dynamic wind directions. Currently, most wind farms employ a greedy control strategy \cite{dar2016windfarm} which focuses on maximizing the power output of individual wind turbines by directly facing the incoming wind. While this approach is straightforward and easy to implement,  it fails to account for the complex wake interactions among turbines \cite{howland2019wind}. A control method designed for maximizing the output of a single turbine does not necessarily lead to optimal performance at the wind farm level. Consequently, developing advanced control strategies that coordinate the collective operation of all turbines, rather than treating them in isolation, has emerged in recent years as an active area of research.

An intuitive approach is to develop accurate analytical models of wind farms and design control strategies based on them. Model-based methods, such as model predictive control \cite{vali2017adjoint,vali2019adjoint,kheirabadi2021real}, have been extensively studied for maximizing power capture. In \cite{campagnolo2020wind}, a  method was developed using a model-based optimization technique to generate a wake-steering look-up table (LUT), which assigns discrete yaw set-points to each turbine depending on the inflow conditions to maximize power generation. However, these methods depend on the accuracy of the underlying wind farm models, whose performance can be affected by inevitable modeling inaccuracies and unmodeled physics. 
To address these limitations, several model-free control algorithms have been proposed, including particle swarm optimization \cite{gionfra2019wind}, genetic algorithms \cite{wang2016novel}, and Bayesian optimization \cite{park2016bayesian}.  While these approaches eliminate the need for an explicit wind farm model, they share a structural limitation: they perform direct optimization on the farm, treating it as a static input-output map. Each candidate solution must be evaluated over long averaging windows to filter out turbulence, and the search must re-converge whenever the inflow changes. These methods therefore deliver point solutions for quasi-static conditions rather than a feedback control policy, which makes them ill-suited to continuously varying ambient conditions. 

In recent years, reinforcement learning (RL) has emerged as a promising approach that has been successfully applied in a wide range of fields, including aerospace \cite{wallace2024reinforcement}, robotics \cite{chai2022design}, and power systems \cite{meng2024online}. In RL, the agent gradually discovers the optimal control policy by continuously learning from its interactions with the environment. Two features distinguish RL from the direct optimization methods discussed above. First, RL learns a state-feedback control policy rather than a point solution: once trained, the policy maps the measured operating conditions to control actions in real time, with no need to re-run an optimization when the inflow changes. Second, RL maximizes a long-term cumulative objective, which naturally accommodates transient behavior and actuation penalties alongside power capture. The effectiveness of RL for wind farm control has been demonstrated in several studies \cite{zhao2020cooperative,dong2023reinforcement,huang2024reinforcement}. In \cite{zhao2020cooperative}, a knowledge-assisted RL algorithm was proposed for thrust coefficient control to enhance power generation under time-varying wind speeds. In \cite{dong2023reinforcement}, an automatic grouping technique was integrated into a multi-agent transfer learning-based RL algorithm to maximize power capture in large-scale wind farms. In \cite{huang2024reinforcement}, a Multi-Agent Proximal Policy Optimization (MAPPO) algorithm was proposed for wind farm control problems with multiple control objectives. Although the above studies achieve notable improvements in power generation compared to the benchmark greedy method, they still have some limitations. First, all these methods rely on online RL, where the agents are trained by interacting frequently with simulators to collect large sets of training samples. This process is computationally demanding, especially when using sophisticated fluid dynamics simulators. Additionally, a `sim-to-real' gap often exists between wind farm simulators and their real-world counterparts. Secondly, most existing RL-based wind farm control methods assume a specific wind direction, whereas dynamic changes in wind direction significantly alter wake interactions among turbines. Varying wind speed presents a simpler challenge for an RL agent, because setpoints change less and more smoothly with respect to wind speed than to wind direction. In contrast, training an RL agent in a setting with changing wind direction is far more challenging, as the agent must adapt to different inter-turbine wake interaction patterns, making the training process more complex. 

A natural way to avoid both the simulator burden and the `sim-to-real' gap is to learn directly from historical data logged by wind farms during routine operation, which modern supervisory control and data acquisition systems already collect in abundance. It is important to recognize, however, that directly utilizing such data to train conventional RL models entails some inherent limitations. On-policy RL methods, such as Proximal Policy Optimization (PPO) \cite{schulman2017proximal}, cannot directly use offline data because they require training samples to be generated using the same control policy that is being learned. Off-policy RL methods, such as Deep Deterministic Policy Gradient (DDPG) \cite{lillicrap2015continuous} and Twin Delayed Deep Deterministic Policy Gradient (TD3) \cite{fujimoto2018addressing}, can in principle leverage offline data, but they often suffer from ``extrapolation errors'', that is, inaccurate value estimates for state-action pairs that are not contained in the dataset, which can negatively impact their performance. 

Motivated by the above limitations of existing methods, this article develops an offline RL framework for wind farm power maximization under dynamic wind directions. The framework retains the policy-learning advantages of RL while replacing all environment interactions with learning from a fixed dataset of logged operation. In this way, it requires no wake model, no plant-in-the-loop probing, and no simulator interaction, which are the respective bottlenecks of the three method families discussed above, while still producing a real-time feedback policy. The main novelties and contributions of this article are as follows:
\begin{enumerate}
    \item At the framework level, an offline RL approach is proposed for farm-level yaw control under dynamic wind directions. In contrast to model-based methods \cite{vali2017adjoint,vali2019adjoint,kheirabadi2021real,campagnolo2020wind}, it requires no wake model at either the training or the deployment stage and uses only sensor-measurable data. In contrast to online RL methods \cite{zhao2020cooperative,dong2023reinforcement,huang2024reinforcement}, it learns entirely from a fixed dataset of logged operation, eliminating simulator interactions and the associated `sim-to-real' gap. As with any offline learning method, the quality of the resulting policy depends on how informative its training data are. In particular, the logged data should contain sufficient variation in yaw settings for the effect of wake steering to be learned. However, these data do not need to come from any particular controller. They may be generated by a look-up table, a model predictive controller, exploratory yaw perturbations during commissioning, or any mixture of operational strategies already deployed on the farm. The framework thereby supports a practical workflow: the farm initially operates under any available wind farm controller; the proposed offline RL controller, once trained from the accumulated operational data, is then deployed; and the data it generates in operation feed subsequent rounds of offline refinement, without any further interaction with the plant for training.
    \item At the algorithmic level, the MTD3-BC method is developed by extending the TD3-BC framework \cite{fujimoto2021minimalist} with a new action consistency regularization term. This term explicitly accounts for the slow response of yaw actuators and suppresses the control chattering that arises when the standard TD3-BC algorithm is applied to this problem. In addition, a wind-direction-conditioned actor architecture and dataset-level state normalization are designed, enabling the measured wind direction to effectively guide policy generation under varying inflow conditions.
    \item At the experimental level, the proposed framework is validated in wind tunnel tests on a farm of three scaled turbines driven by wind direction time series recorded in the field. To the best of our knowledge, this is the first experimental validation of an offline RL-based wind farm control strategy. The learned policy performs on par with a data-calibrated LUT benchmark and an online PPO controller, marginally outperforms the best behavior policy in its training dataset, and requires only approximately 5\% of the training cost of PPO. Three successive deploy-and-retrain iterations further show the practicality of the offline refinement workflow.
\end{enumerate}

In the remaining sections, the wind farm control problem is formulated in Section~\ref{se2}. The modified TD3-BC algorithm is detailed in Section~\ref{se3}. To validate its effectiveness, the results of wind tunnel tests under varying wind directions are presented and discussed in Section~\ref{se4}. Finally, we conclude the article in Section~\ref{se5}.

\section{Problem Formulation}
\label{se2}
Consider a wind farm with $n$ turbines that are denoted by $\mathcal{WT}_1,\mathcal{WT}_2,\cdots,\mathcal{WT}_n $. Generally, for a given turbine-level torque and pitch control law, the power $P_i$ generated by a turbine $\mathcal{WT}_i$ is related to the inflow speed $U_i$ at its location, the wind direction $\Phi_i$, and the yaw offset $\gamma_i$, which yields
\begin{equation}
    \label{eq1}
    P_i = h(U_i,\Phi_i,\gamma_i).
\end{equation}
From (\ref{eq1}), it is evident that the total power output $P = \sum_{i=1}^{n} P_i $ can be controlled by adjusting the yaw angles of individual turbines in the wind farm. However, the function
$h$ governing this relationship is highly complex and difficult to derive accurately. In the RL-based method proposed in the following sections, we adopt a data-driven, model-free approach, which allows the function
$h$ to remain entirely unknown.

In this article, our primary objective is to maximize the power generation of a wind farm by controlling the yaw angles of individual turbines. However, from an engineering perspective, large yaw misalignments can impose significant structural loads, potentially shortening the lifespan of the turbines. Therefore, our control strategy aims to balance the two objectives. Based on this trade-off, we define the following instantaneous reward function:
    \begin{equation}
    \label{reward}
        r(t) = \sum_{i=1}^{n}P_i(t)-k\sum_{i=1}^{n}\left|\gamma_i(t)\right|
    \end{equation}
where $k>0$ is  a user-defined weight between power boosting and yaw misalignment penalization.
    
Thus, the goal can be formulated as an optimal control problem aimed at maximizing the expected cumulative discounted reward:
\begin{equation}
\label{objective}
J(\pi) = \mathbb{E}_{\pi}\!\left[ \sum_{t=0}^{\infty} \xi^{t}\, r(t) \right],
\end{equation}
subject to
\begin{equation}
  - \gamma _{\max} \leq\gamma _i(t) \leq \gamma _{\max},
\end{equation}
where $ 0<\xi < 1 $ is the discount factor and $ \gamma_{\max}$ is the upper bound on the yaw angles. 

It is important to note that wind farm control is a highly complex problem. A greedy approach, where each turbine's yaw angle is adjusted to directly face the incoming wind for local power maximization, may initially appear optimal; however, as discussed above, it ignores the wake coupling among turbines and is therefore generally suboptimal at the farm level. The complexity is further exacerbated when dynamic wind direction changes are considered. In fact, when the wind direction changes, the wake interaction pattern among the turbines changes as well, introducing significant variability into the behavior of the system. This dynamic environment adds considerable difficulty to the learning process, as the RL agent must adapt to varying wake interactions and adjust its control strategy accordingly. In the following section, we will present a novel offline RL-based control framework designed to tackle this problem.

\section{Modified TD3-BC Algorithm for Wind Farm Control}
\label{se3}
\subsection{Offline Dataset}
\label{dc}
We consider a standard Markov Decision Process (MDP) setting $\left(\mathcal{S},\mathcal{A},\mathcal{R},p,\xi\right)$, where $\mathcal{S}$ is the state space,  $\mathcal{A}$ represents the action space, $\mathcal{R}:\mathcal{S}\times\mathcal{A}\to\mathbb{R}$ is the reward function, $p$ denotes the transition dynamics, and $\xi\in(0,1)$ is the discount factor. The objective of the RL agent is to find a deterministic policy $\pi:\mathcal{S}\to\mathcal{A}$ that maximizes the
expected return $J(\pi)$ defined in \eqref{objective}. 

In the offline RL setting, the agent learns a policy $\pi$ from a fixed dataset $ \mathcal{D} $ consisting of single-step transitions of the form  $ (s_t,a_t,r_t,s_{t+1}), s_t, s_{t+1} \in \mathcal{S}, a_t \in \mathcal{A},
r_t = \mathcal{R}(s_t, a_t)$. Unlike online RL, the offline RL agent does not need to interact directly with the environment, allowing it to fully utilize previously logged data. This approach is particularly beneficial in scenarios where real-time interaction with the environment is impractical, risky, or costly. 

In this work, the offline wind farm dataset $ \mathcal{D} $ comprises single-step transitions $ (s_t,a_t,r_t,s_{t+1}) $ collected under different control algorithms. The reward function has been defined in (\ref{reward}). The state and action contained in this dataset are defined as follows:
\begin{itemize}
	\item State: The state $ s_t$ of a wind farm includes both internal and external status information. The internal state reflects the operational conditions of the wind turbines, including variables such as yaw angles. The external state reflects environmental factors affecting turbine performance, such as measured wind direction, wind speed at each turbine location, and turbulence intensity.
	
	\item  Action: The action $ a_t$ represents the yaw references that wind turbines should follow.
\end{itemize}

\subsection{Modified TD3-BC Algorithm}
TD3-BC is a state-of-the-art offline RL method, which minimally modifies the standard online off-policy TD3 method by integrating a simple BC regularization term into the policy update step. This modification pushes the policy towards prioritizing actions that are well-represented in the observed dataset, thus mitigating the extrapolation error caused by out-of-distribution actions \cite{fujimoto2021minimalist}.

In the TD3-BC framework, like the standard TD3 algorithm, a total of six neural networks are typically employed under the actor-critic architecture. Specifically, two critic networks, denoted as $Q_{\theta_1}$ with parameter $\theta_{1}$ and $Q_{\theta_2}$ with parameter $\theta_{2}$, are used to approximate the action-value function $Q\left(s,a\right) $, which estimates the expected long-term reward from a given state $s$ after executing an action $a$. At timestep $t$, for a given policy $\pi$, the action-value function satisfies the following equation: 
\begin{equation}
\label{q-value}
   Q_{\pi}(s_t, a_t) = \mathbb{E}_{s_{t+1} \sim p} [r_t + \xi Q_{\pi}(s_{t+1}, \pi(s_{t+1}))].
\end{equation}

The actor network $\pi_{\phi}(\cdot)$ with parameter $\phi$ is responsible for learning the control policy. The primary goal of the policy update in TD3-BC is to find a high-performing policy $ \pi^{\star}(s) $ by maximizing the expected Q-value function defined in (\ref{q-value}).

To improve training stability, each actor and critic network is paired with a corresponding target network: $Q_{\theta^{\prime}_1}$, $Q_{\theta^{\prime}_2}$, and $\pi_{\phi'}$. These target networks help stabilize the learning process and mitigate fluctuations during training. The parameters of the target critic and actor networks, denoted as  $\theta'_1$, $ 	\theta'_2$ and $ \phi'$, are updated using a soft replacement strategy to gradually track their main counterparts:
\begin{equation}
	\label{soft} 
	\begin{split} 
		&\theta'_j \leftarrow \tau \theta_j + (1 - \tau) \theta'_j, \; j=1,2,
		\\ 
		&\phi' \leftarrow \tau \phi + (1 - \tau) \phi',
	\end{split}
\end{equation}
where $ \tau \in (0,1] $ is the soft update factor, controlling the speed of adaptation.

The main critic networks are updated using the following temporal-difference (TD) error-based loss functions:
\begin{equation}
\label{td3_critic}
    L(\theta_j) = \frac{1}{N}\sum_{i=1}^{N}\left(y_i-Q_{\theta_j} (s_i,a_i)\right)^2,
\end{equation}
\begin{equation}
\label{td_error}
    y_i =  r_i+\xi \min_{j=1,2}Q_{\theta^\prime_j}\left(s_{i+1},\pi_{\phi'}\left(s_{i+1}\right)+\epsilon\right),
\end{equation}
where $N>0$ is the size of a batch $\mathcal{B}$ sampled in each iteration. The index $i$ denotes a generic sample in the batch, temporarily replacing the timestep index $t$. As indicated in \cite{fujimoto2021minimalist}, to prevent overfitting, noise $\epsilon$ is added to the target policy output, where $\epsilon = \operatorname{clip}(\tilde{\epsilon}, -\bar{c}, \bar{c}),
\tilde{\epsilon} \sim \mathcal{N}(0, \sigma_1^{2})$. Here, $\mathcal{N}$ denotes the normal distribution, $\sigma_1$ the noise scale, and $\bar{c}$ the clipping range.

The policy learning objective function to be maximized is given in (\ref{td3_policy}). This objective function consists of three terms, one of which is a novel action consistency regularization term introduced in this work. The motivation for incorporating this term arises from the observation that directly applying the TD3-BC control algorithm can lead to sensitivity issues: small variations in the input state may cause large variations in the output action, resulting in a chattering control behavior. Moreover, given the relatively slow response of yaw actuators, rapid changes in the yaw reference may prevent the physical system from accurately following the desired yaw commands. This mismatch can further degrade the control performance. 

Specifically, the first term corresponds to the core reinforcement learning objective, which aims to maximize Q-values and thus encourage actions with high rewards. The second term is the behavioral cloning (BC) component, which encourages the agent to favor policies that stay close to those in the provided training dataset. The third term enforces action consistency by penalizing discrepancies between actions generated for similar inputs, thereby mitigating the chattering effect and promoting smoother control actions.  The objective function is written as
\begin{figure*}[t]
  \centering
  \includegraphics[width=4.7in]{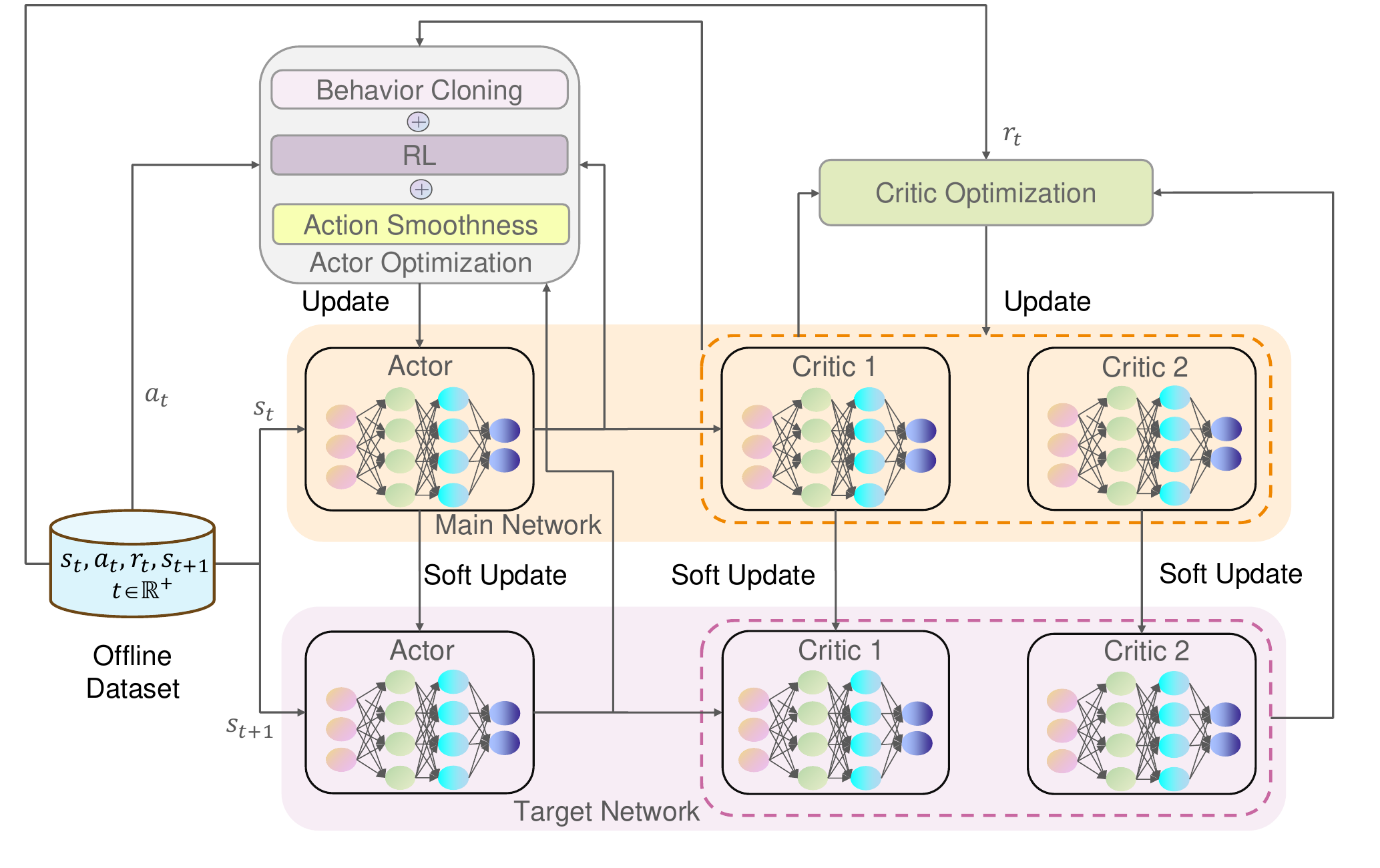} 
  \caption{Schematic representation of MTD3-BC.}
  \label{structure}
\end{figure*}
\begin{equation}
\label{td3_policy}
\begin{split}
\mathcal{J}(\phi) = \frac{1}{N}\sum_{i=1}^{N}\bigg[
&\underbrace{\lambda_r Q_{\theta_1}\big(s_i, \pi_\phi(s_i)\big)}
_{\text{Q-value maximization}}
-\underbrace{\big\|\pi_\phi(s_i) - a_i\big\|^2}_{\text{Behavior cloning}}\\
&-\underbrace{\lambda_s\big\|\pi_\phi(s_i)-\pi_\phi(\tilde{s}_i)\big\|^2}
_{\text{Action consistency regularization}}\bigg],
\end{split}
\end{equation}
where $\lambda_{s}>0$ is a user-defined hyperparameter controlling the strength of the action-consistency regularization. $\tilde s_{i}$ is sampled from a Gaussian distribution centered at the original state $s_{i}$, satisfying $\tilde s_{i}\sim \mathcal{N}(s_i, \sigma_2^2 I) $, where $ \sigma_2$ is the state perturbation scale. Moreover, a hyperparameter $ \lambda_r $ is introduced to balance the RL and BC components by scaling the Q-value term. For a sampled mini-batch of $ N $ transitions $ (s_i, a_i) $, the scalar $ \lambda_r $ is defined as
\begin{equation}
	\lambda_r = \frac{\alpha}{\dfrac{1}{N}\sum_{i=1}^{N}
\big|Q_{\theta_1}(s_i, a_i)\big|},
\end{equation}
where $ \alpha >0$ is a user-defined RL-BC trade-off coefficient. The schematic representation of the proposed modified TD3-BC wind farm control framework is shown in Fig.~\ref{structure}. 

To improve training stability, the modified TD3-BC normalizes all states in the dataset once before training: 
\begin{equation}
\label{norm}
 s_{i} \leftarrow \frac{s_{i}-\mu}{\sigma +\varepsilon},
\end{equation}
where $ \mu $ and $ \sigma $ are the mean and standard deviation for each state dimension, respectively, and $  \varepsilon = 10^{-3}$ is a small constant. State normalization is a standard technique in many deep learning algorithms and is particularly suitable for offline RL settings, as the dataset remains fixed throughout one training process. 

The training process of MTD3-BC also incorporates a delayed updating strategy, where the actor and target networks are updated once every $d$ critic updates. Based on the above description, the full training details of modified TD3-BC (MTD3-BC) are presented in Algorithm~\ref{alg:TD3-BC}.  

\begin{algorithm}
    \caption{Modified TD3-BC (MTD3-BC)}
    \label{alg:TD3-BC}
    Compute the mean $\mu$ and standard deviation $\sigma$ of each state dimension over the offline dataset $\mathcal{D}$, and normalize all states in $\mathcal{D}$ using (\ref{norm})\;
    Initialize main critic networks $Q_{\theta_1}$, $Q_{\theta_2}$, and main actor network $\pi_{\phi}$ with random parameters $\theta_1$, $\theta_2$, $\phi$\;
    Initialize the target networks with $\theta'_1 \gets \theta_1$, $\theta'_2 \gets \theta_2$, $\phi' \gets \phi$; set the total number of training steps $T_{\max}$\;
    \For{$t = 1$ to $T_{\max}$}{
        Randomly sample a batch of $N$ transitions $(s_i, a_i, r_i, s_{i+1})$ from the offline dataset $\mathcal{D}$\;
        Compute target actions with target policy smoothing noise $a_{i+1} = \pi_{\phi'}(s_{i+1}) + \epsilon$, $\epsilon = \operatorname{clip}(\tilde{\epsilon}, -\bar{c}, \bar{c}),\tilde{\epsilon} \sim \mathcal{N}(0, \sigma_1^{2})$\;
        Compute perturbed states $\tilde s_{i}\sim \mathcal{N}(s_i, \sigma_2^2 I)$\;
        Update the parameters $\theta_1$, $\theta_2$ of the main critic networks using (\ref{td3_critic}) and (\ref{td_error})\;
        \If{$t\mod d = 0$}{
            Update the parameter $\phi$ of the main actor network by (\ref{td3_policy})\;
            Update parameters $\theta'_1$, $\theta'_2$, $\phi'$ of the target networks via the soft replacement technique by (\ref{soft})\;
        }
    }
\end{algorithm}


\section{Experimental Validation}
\label{se4}
\subsection{Experimental Setting}
\begin{figure}[!t]
  \begin{center}
  \includegraphics[width=3.2in]{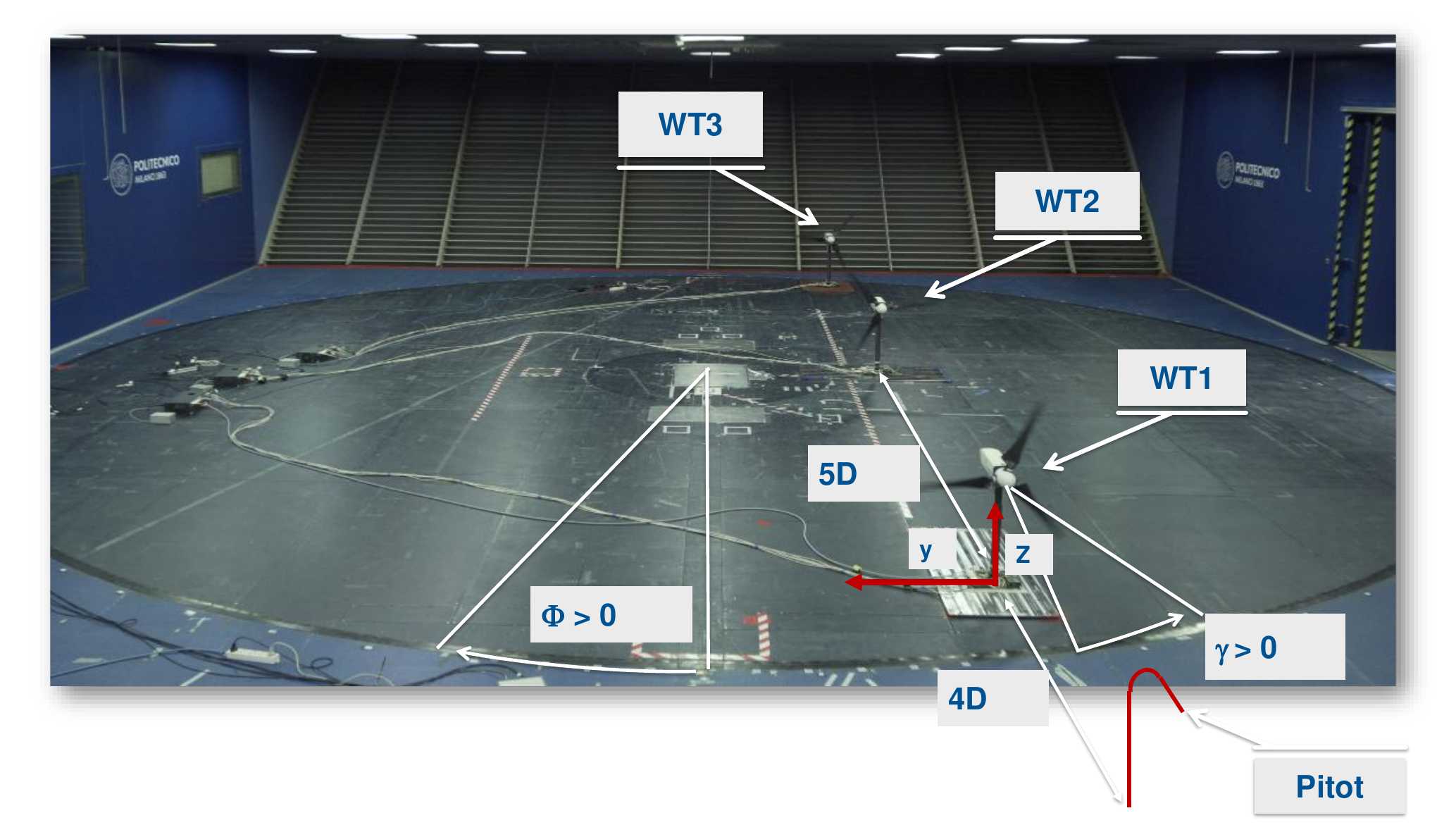}
  \caption{Experimental setup.}\label{setup}
  \end{center}
\end{figure}
\begin{figure}[!t]
    \centering
    \includegraphics[width=3.1in]{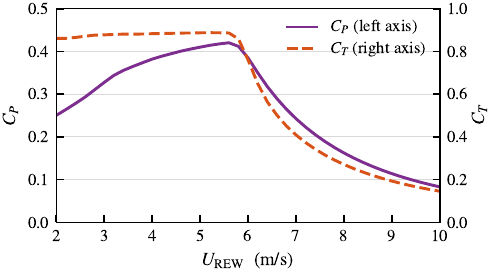}
    \caption{Power coefficient $C_P$ (left axis) and thrust coefficient $C_T$ (right axis) curves with respect to the wind speed.}
    \label{powercurve}
\end{figure}

The experimental setup is depicted in Fig.~\ref{setup}. A group of three scaled G1 wind turbines \cite{bottasso2022wind}, denoted as WT1 (upstream), WT2 (center), and WT3 (downstream), is mounted on the 13-m-diameter turntable within the atmospheric test section of the Politecnico di Milano wind tunnel \cite{bottasso2014wind}. Each turbine model has a rotor diameter, hub height, and rated rotor speed of $1.1\ \mathrm{m}$, $0.825\ \mathrm{m}$, and $850\ \mathrm{rpm}$, respectively. Wind turbines are spaced 5 rotor diameters ($5D$) apart longitudinally and positioned $1.5D$ to the left of the centerline of the turntable when viewed upstream \cite{campagnolo2020wind,bottasso2014wind,campagnolo2022further}. The power coefficient $C_P$ and thrust coefficient $C_T$ as functions of the rotor-equivalent wind speed $U_\mathrm{REW}$ are shown in Fig.~\ref{powercurve}. The rated wind speed is $U_\mathrm{rated}=5.7 \ \mathrm{m/s}$. Below the rated wind speed (Region II), torque control is implemented, whereas above $U_\mathrm{rated}$ blade pitch control is applied (Region III). A standard controller is integrated with the turbines to realize the aforementioned torque and pitch control. 

During the experiment, the turntable can be rotated to simulate low-frequency, large-scale variations in dynamic wind direction. The turntable angle, denoted as $\Phi$, is null when the imaginary line connecting the turbine row is aligned with the  central axis of the wind tunnel. A positive $\Phi$ corresponds to a clockwise rotation of the turntable when observed from above, while a positive yaw angle $\gamma$ represents an anti-clockwise rotation, as illustrated in Fig.~\ref{setup}. Based on this definition, a positive wind direction $\Phi$ indicates wind coming from the left side of the turbine row when looking in the upstream direction. Furthermore, a time speed-up factor of 80 is applied to achieve temporal scaling relative to the full-scale system, meaning that 1 hour of wind tunnel testing corresponds to approximately 3.3 days in the field. Accordingly, dynamic wind direction measurements over five days, recorded at 1 Hz at an onshore test site in northern Germany, are compressed into 90 minutes of wind data for use in the experiment, with directional variations within a range of $\pm 15^\circ$. 

To enable the periodic calibration of the wind tunnel and turbine sensors, the 90-minute wind record was divided into nine wind profiles, each lasting 10 minutes and 10 seconds. To process the recorded data, the first 10 seconds of each $610 \ \mathrm{s}$ profile were excluded. Sensor zeroing was performed both before and after each $610 \ \mathrm{s}$ segment to mitigate the impact of zero drift on the measured data.

In this wind tunnel test, the maximum allowable yaw misalignment is $ \gamma_{\max} = 30^\circ$. The inflow turbulence intensity is set to $\mathrm{TI}=6\%$. The wind speed at the Pitot tube, which is installed $4D$ upstream of WT1 at hub height, is $U=5.35 \ \mathrm{m/s}$.  

\subsection{Dataset Creation, Network Architecture, and Hyperparameter Setting}
In the wind tunnel test, we obtained the following signals: the filtered turntable encoder signal, representing the measured turntable angle $\Phi_{m}$, and nacelle orientation $\gamma_{ni}$ of each turbine. As the wind direction $\Phi_{i}$ at each turbine is unavailable, we substituted it with the measured global wind direction $\Phi_{m}$ for the entire farm. Next, the individual measured turbine yaw offset relative to the wind, $\gamma_i$, was  computed as 
\begin{equation}
     \gamma_i =  \gamma_{ni} - \Phi_{m}.
\end{equation}

In addition, the rotor speed $n_i$ (in rpm) and generator torque $T_i$ were measured for each turbine. The power (in kW) generated by the $i$-th turbine was then computed as 
\begin{equation}
\label{power_cal}
    P_i =
\frac{T_i n_i \pi}{30000}.
\end{equation}

In this experiment, the state $s_t$ stored in the dataset was obtained by concatenating the yaw angles of individual turbines relative to the wind direction, together with the wind direction itself, i.e., $s_t = \left[\gamma_1, \gamma_2, \gamma_3, \Phi_m \right]^{\mathrm{T}}$. The action $a_t$ corresponds to the yaw misalignment command applied to the wind farm. The reward $r_t$ was computed according to (\ref{reward}) and (\ref{power_cal}). The state-action-reward tuples were sampled at 1~Hz from raw experimental measurement data.
\begin{figure}[!t]
  \begin{center}
  \includegraphics[width=0.9\columnwidth]{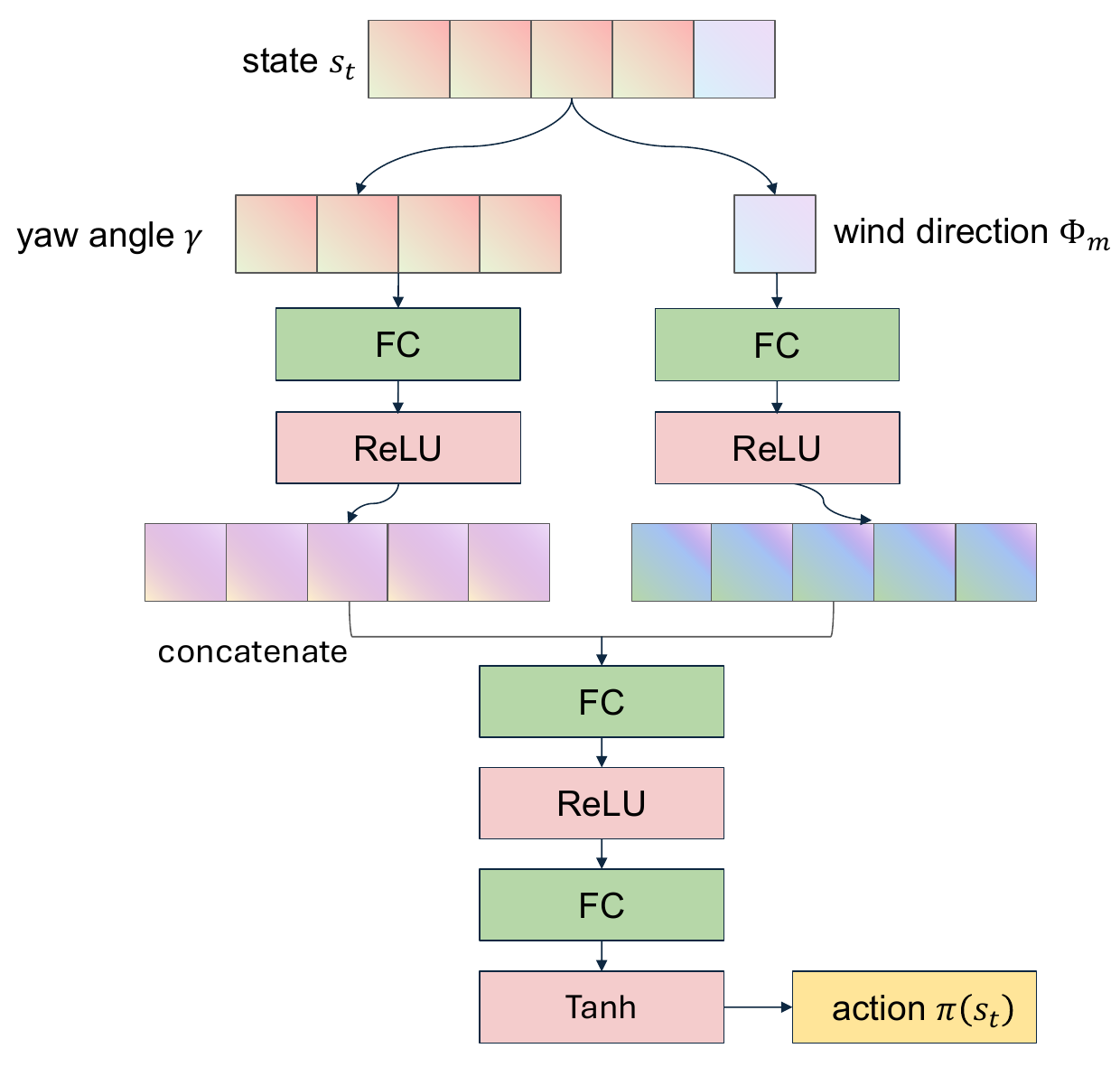}
  \caption{Architecture of actor network.}\label{actor}
  \end{center}
\end{figure}

The initially provided offline wind farm training dataset, $\mathcal{D} $, consisting of $  (s_t,a_t,r_t,s_{t+1})$ samples, was derived from 80~GB of results obtained by applying the greedy method and the LUT method to wind tunnel tests under different wind profiles. For the LUT method, three approaches were used, each based on a wind farm model of different fidelity: the FLORIS-based LUT (lower-fidelity model), the data-augmented FLORIS-based LUT (medium-fidelity model), and the data-driven LUT (higher-fidelity model) \cite{campagnolo2020wind}. Since the wind speed and turbulence intensity remained constant in the wind tunnel experiment, the LUTs were computed solely based on wind direction. Additionally, wind direction uncertainties were incorporated into each LUT calculation, modeled as a zero-mean Gaussian distribution with standard deviation $\Phi_\sigma \in \{0^\circ, 2^\circ, 4^\circ, 6^\circ\}$. All the aforementioned LUTs were tested under all nine wind profiles, and the resulting data were processed and included in the offline dataset.

At the beginning of the experiment, the proposed offline RL policy was
trained on the initial dataset and deployed on the turbines. The resulting measurements were then collected, processed and added to the offline dataset for subsequent fine-tuning. This process was repeated three times, with the control algorithm used in each iteration sequentially labeled `Offline RL 1', `Offline RL 2', and `Offline RL 3'. In the following, `LUT' refers specifically to the data-driven LUT ($\Phi_\sigma = 6^\circ)$.

\begin{table}[!t]
    \centering
    \caption{Hyperparameters for offline training.}
    \label{tab}
    \begin{tabular}{l c}
        \toprule
       Parameter symbol & Value \\
        \midrule
        RL-BC trade-off coefficient ${\alpha}$ & 2.5  \\
        Action consistency regularization coefficient  $\lambda_{s}$ & 0.5\\
        Soft update factor ${\tau}$ & 0.005 \\
        Clip range $\bar c$& 0.5\\
        Noise scale $\sigma_1$,$\sigma_2$ & 0.2, 0.2\\
        Discount factor $\xi$ & 0.99 \\
        Learning rate of actor and critic, $l_a, l_c$ & 0.0003, 0.0003 \\
        Size of $\mathcal{B}$ &256 \\
        Network neurons & 256 \\
        Network layers & 3 \\
        Reward weight $k$ & 0.1\\
        Delayed updating steps $d$ & 2 \\
        \bottomrule
    \end{tabular}
\end{table}

Additionally, we observed that in designing the neural network architecture, feeding the normalized wind direction $ \Phi_{m}$ separately into the input layer of the actor network rather than concatenating it with the normalized yaw angle vector $\left[\gamma_1, \gamma_2, \gamma_3\right]^{\mathrm{T}}$ provides ancillary conditioning information and improves the learning process. Specifically, Fig.~\ref{actor} illustrates the customized architecture of the actor network in our modified TD3-BC algorithm.  The hyperparameter values are reported in Table~\ref{tab}. 

\subsection{Experimental Results}

\begin{figure}[!t]
  \begin{center}
  \includegraphics[width=3.1in]{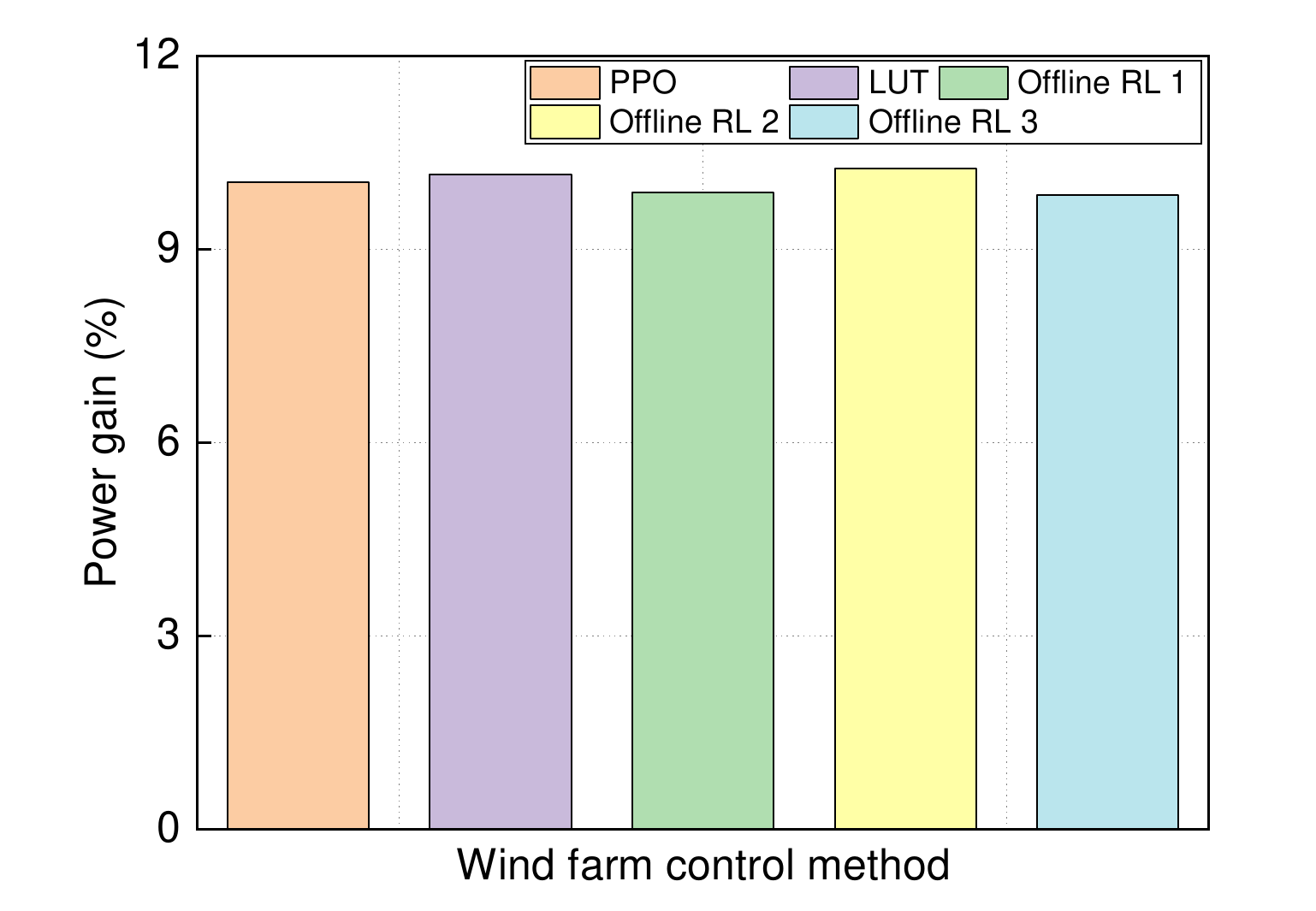}\\
  \caption{Averaged power gains of different control strategies for the wind farm  with respect to the greedy method.}\label{bar}
  \end{center}
\end{figure}

In Fig.~\ref{bar}, we present a bar chart comparing the farm-level average power gains of different control strategies against the greedy policy over the full 90-minute wind direction time series. The results indicate that all three offline RL policies yield a total power increase of approximately 10\% on average compared to the greedy policy. Notably, the best-performing offline RL policy, `Offline RL 2', achieved marginally higher power gains than both the LUT control strategy and the PPO strategy, although the differences among the top-performing methods are of the same order as the variability observed across the three offline RL iterations. These results confirm that the proposed MTD3-BC method matches the performance of a data-calibrated model-based benchmark without requiring any wake model. Moreover, the proposed offline RL policy achieves this level of performance while entirely eliminating the need for direct interaction with the environment or a simulator, significantly reducing computational cost and training time. For the remainder of the discussion, `MTD3-BC' refers specifically to `Offline RL 2'.
\begin{figure*}[!t]
    \centering
    \subfloat[WF]{%
        \includegraphics[width=2.8in]{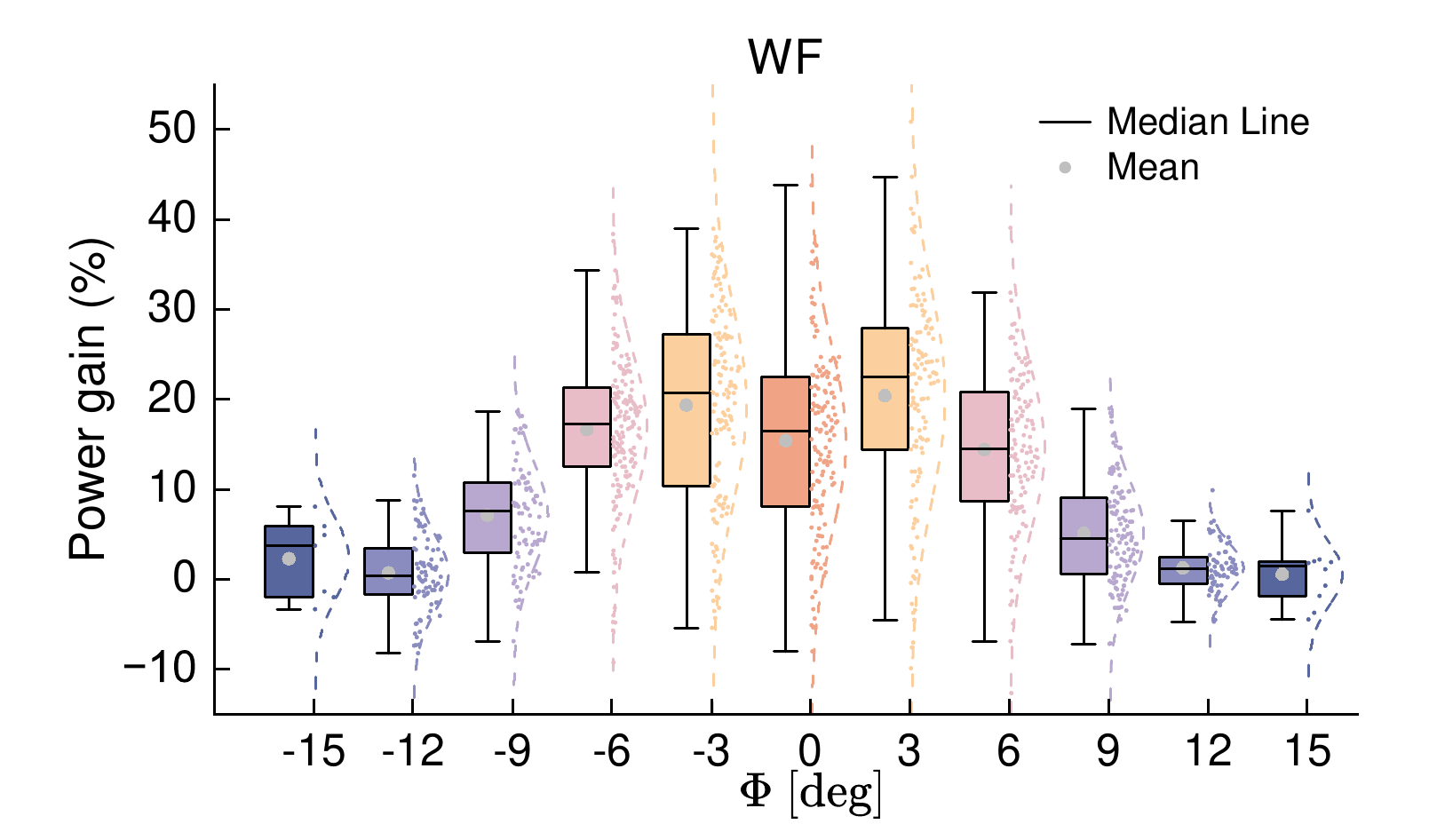}}
    \subfloat[WT1]{%
        \includegraphics[width=2.8in]{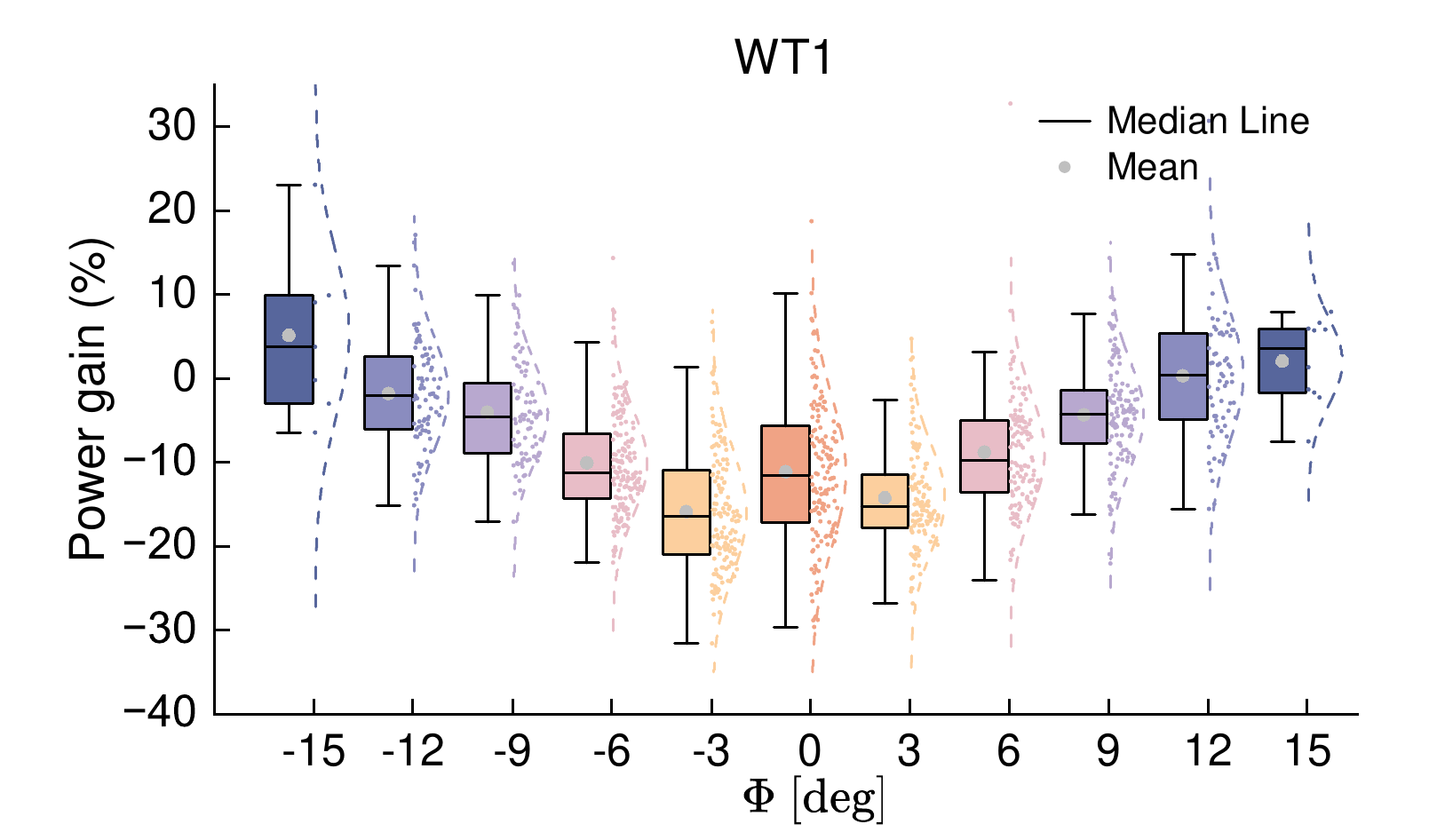}} \par
    \subfloat[WT2]{%
        \includegraphics[width=2.8in]{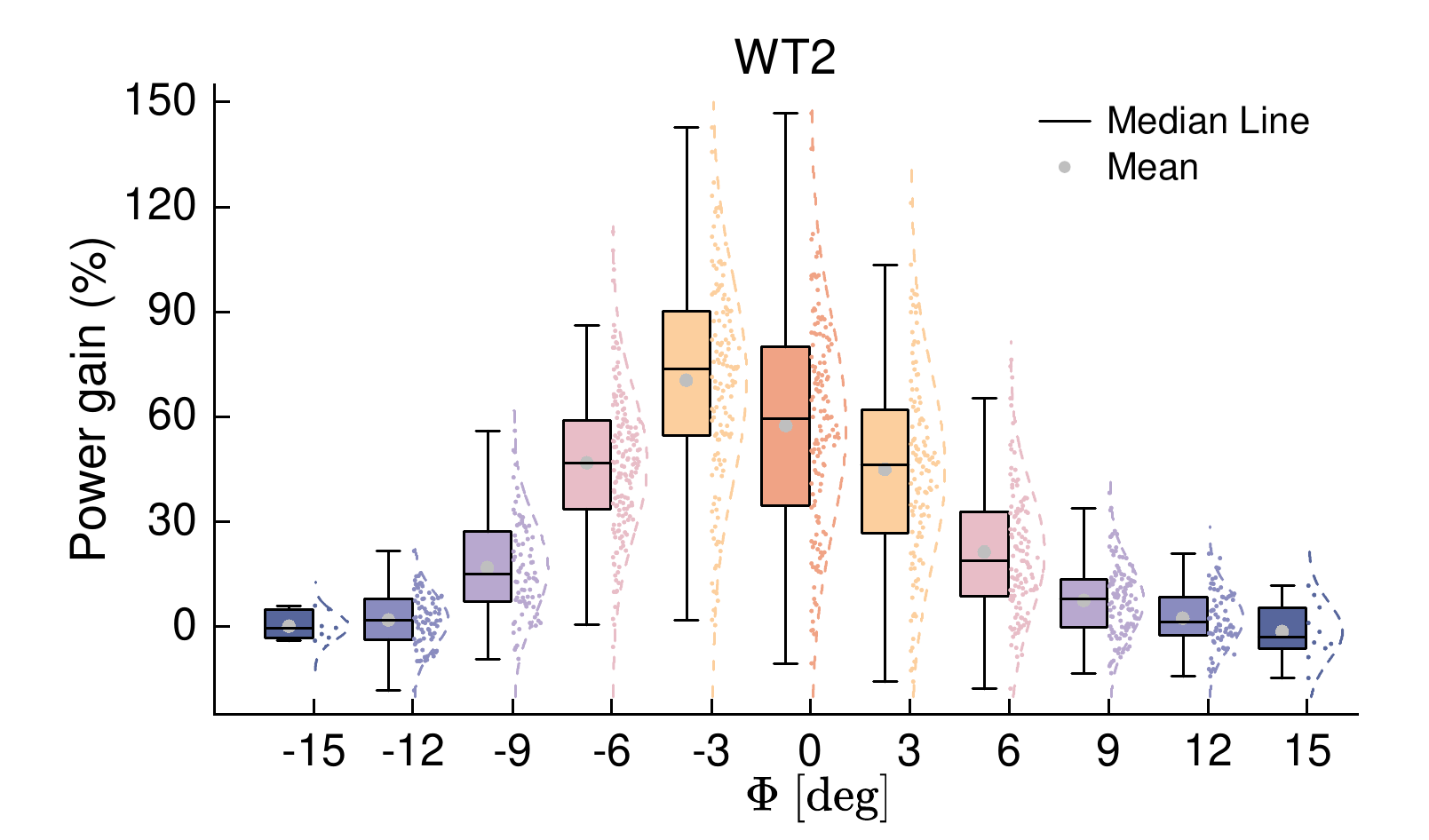}}
    \subfloat[WT3]{%
        \includegraphics[width=2.8in]{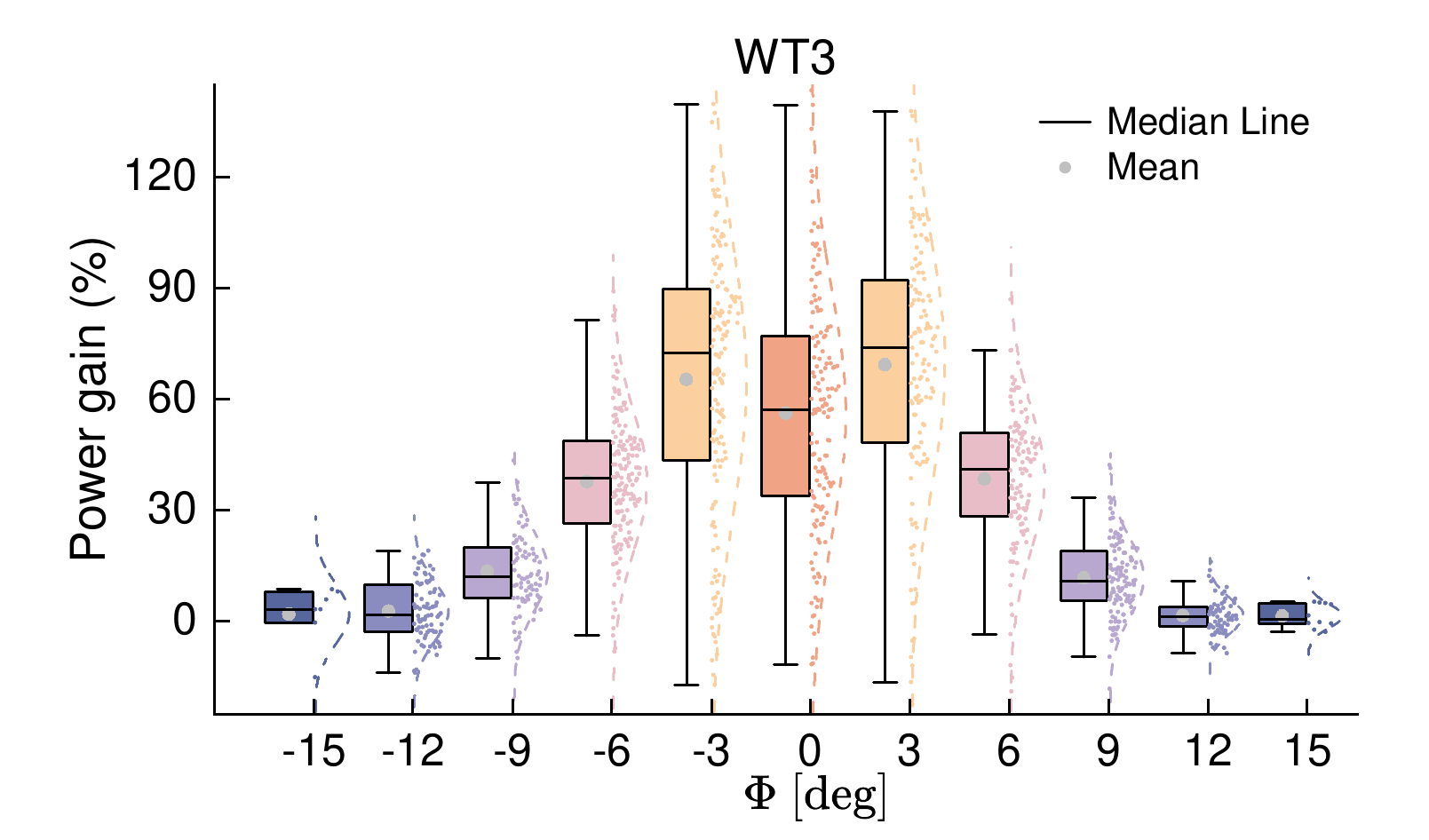}}
    \caption{Farm-level and turbine-level power gains of MTD3-BC relative to the greedy policy under different wind directions.}
    \label{box}
\end{figure*}

Fig.~\ref{box} presents a comparative analysis of power gains achieved by the MTD3-BC control strategy relative to the greedy policy at both the wind farm (WF) and individual wind turbine (WT) levels under varying wind directions. On average, the offline RL strategy consistently outperforms the greedy policy for most wind directions, particularly around $-6^\circ$ to $6^\circ$, where power gains are the highest. For WT1, the average power generated by the MTD3-BC method is lower than that under the greedy strategy in most directions, while WT2 and WT3 exhibit substantial power gains, particularly in the range of $-9^\circ$ to $6^\circ$, where improvements exceed 100\% in some cases. The results demonstrate that the offline RL agent effectively optimizes the overall wind farm performance by adjusting yaw angles in response to wind direction, to redirect upstream wakes. While upstream turbine WT1 may generate less power, the additional power generated by downstream turbines (WT2 and WT3) more than compensates for this, resulting in a net power gain for the entire wind farm.

\begin{figure*}[!t]
    \centering
    \subfloat[WF\label{td3_lut_wf}]{%
        \includegraphics[width=2.8in]{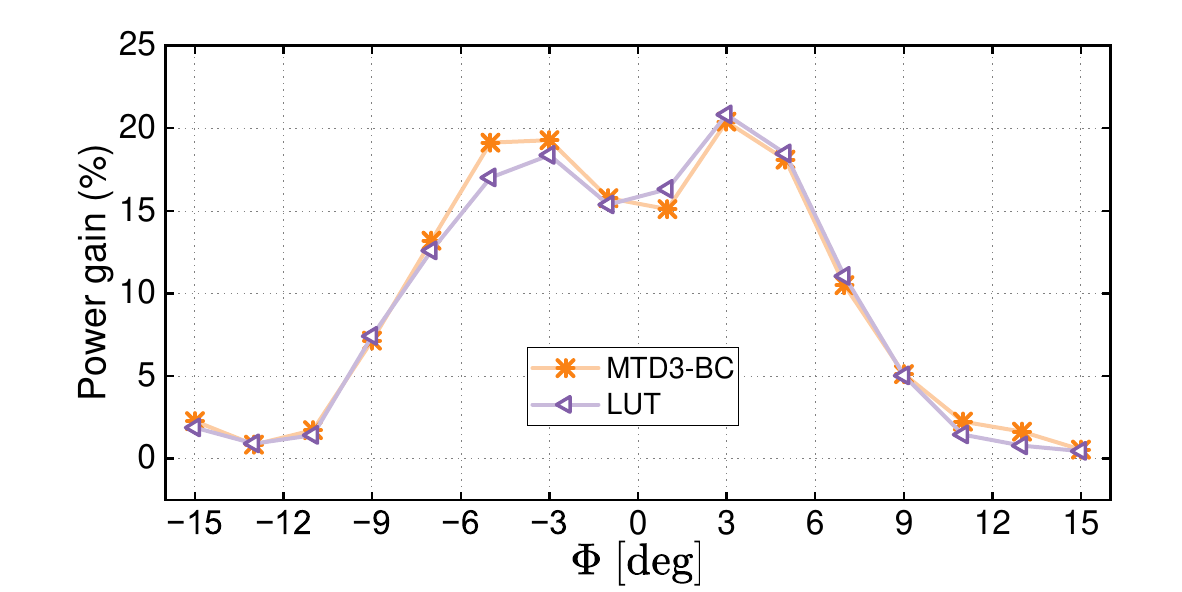}}
    \subfloat[WT1\label{td3_lut_wt1}]{%
        \includegraphics[width=2.8in]{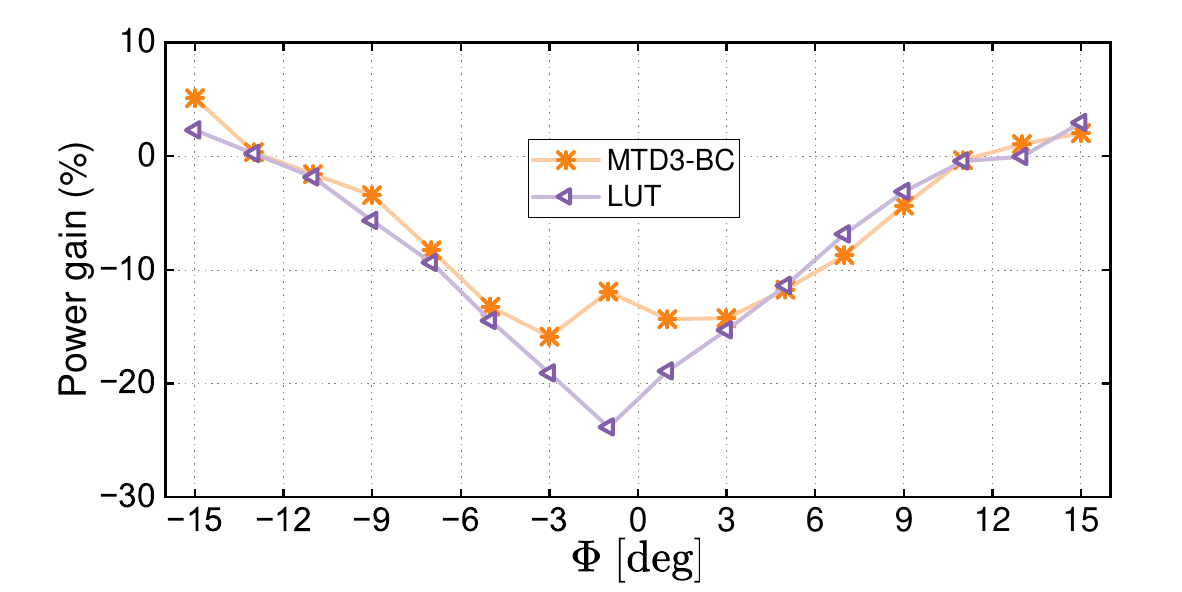}} \par
    \subfloat[WT2\label{td3_lut_wt2}]{%
        \includegraphics[width=2.8in]{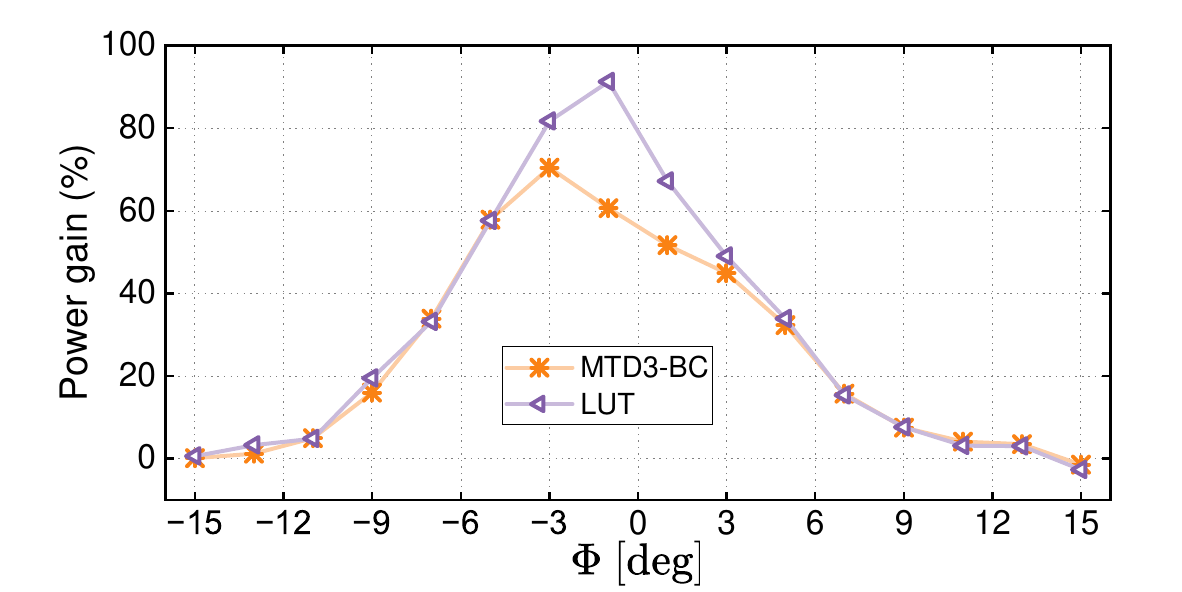}}
    \subfloat[WT3\label{td3_lut_wt3}]{%
        \includegraphics[width=2.8in]{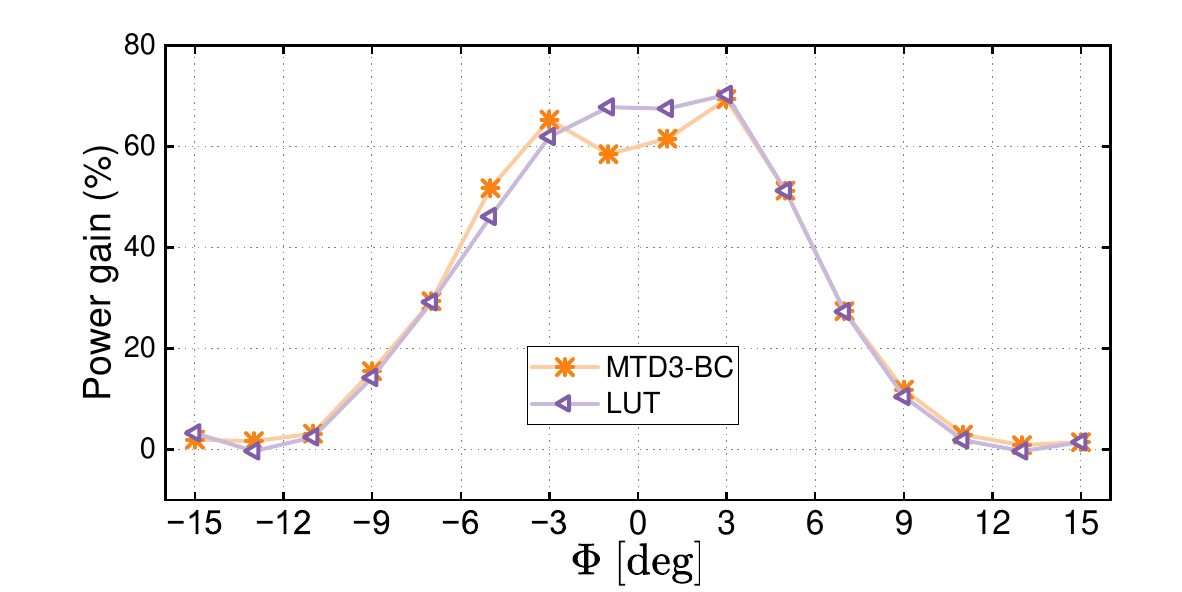}}
    \caption{Farm-level and turbine-level power gains of MTD3-BC and LUT policies relative to the greedy policy under different wind directions.}
    \label{td3_lut_power}
\end{figure*}
\begin{figure*}[!t]
    \centering
    \subfloat[WT1]{
        \includegraphics[width=2.35in]{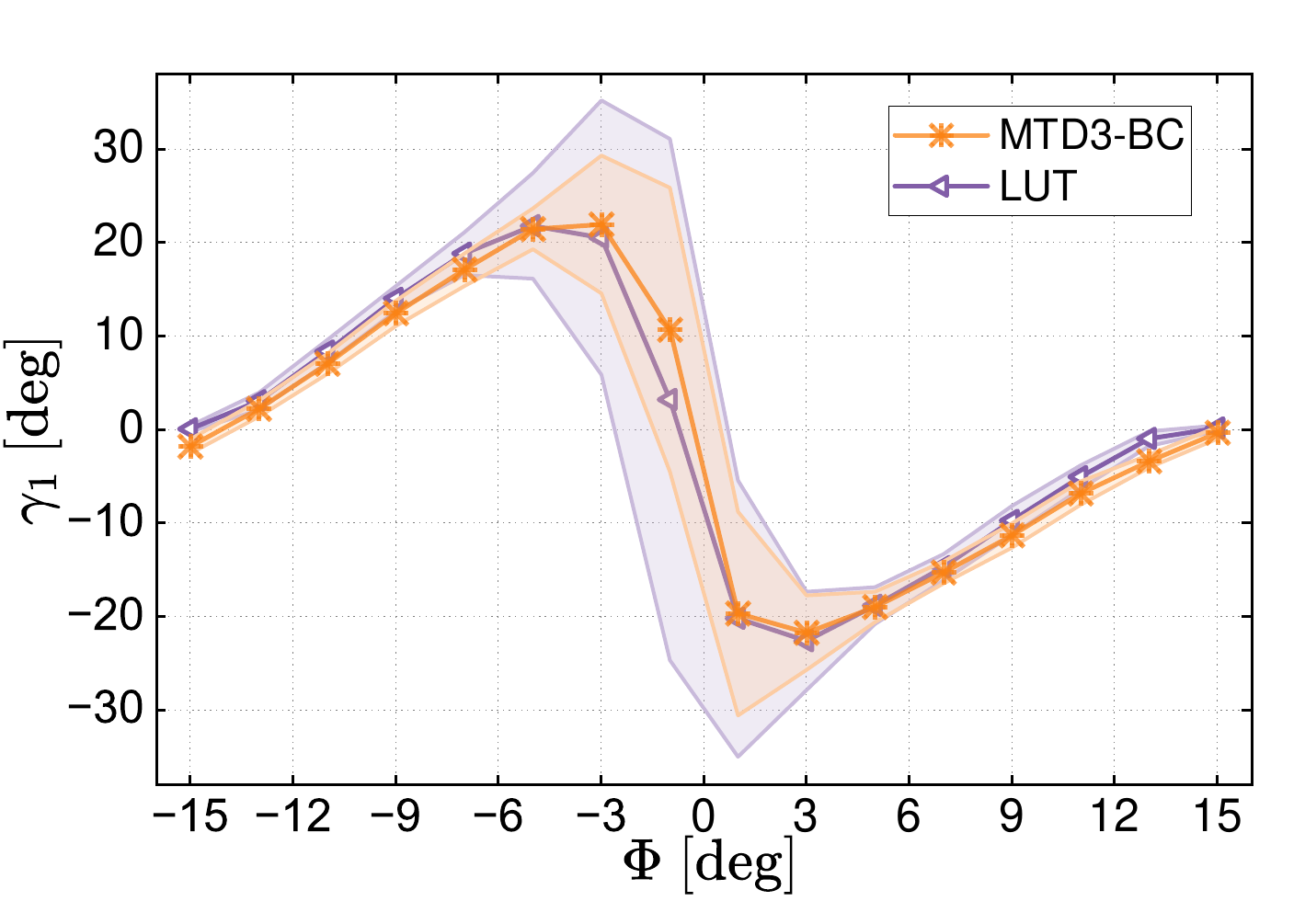}}
    \subfloat[WT2]{
        \includegraphics[width=2.35in]{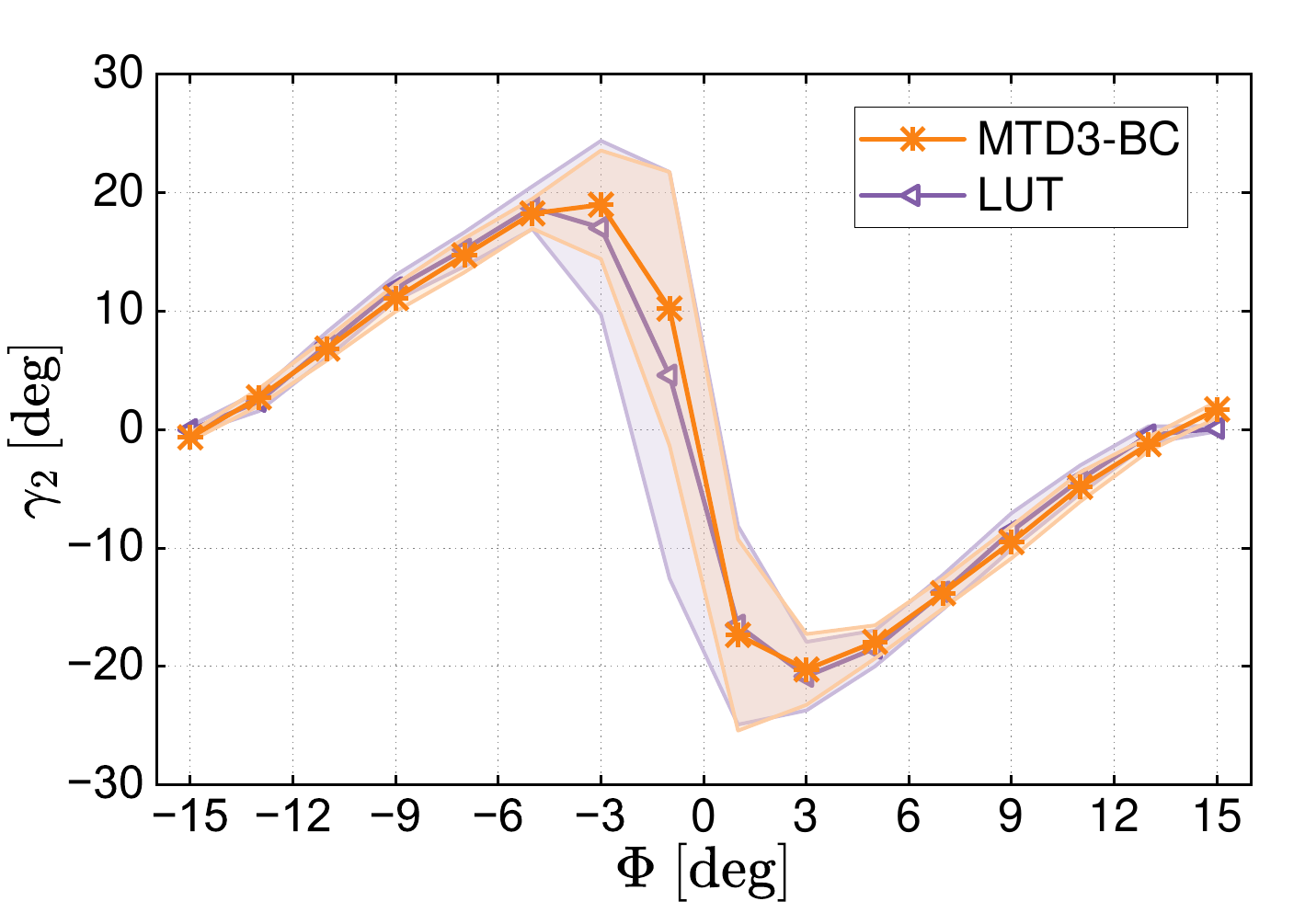}}
    \subfloat[WT3]{
        \includegraphics[width=2.35in]{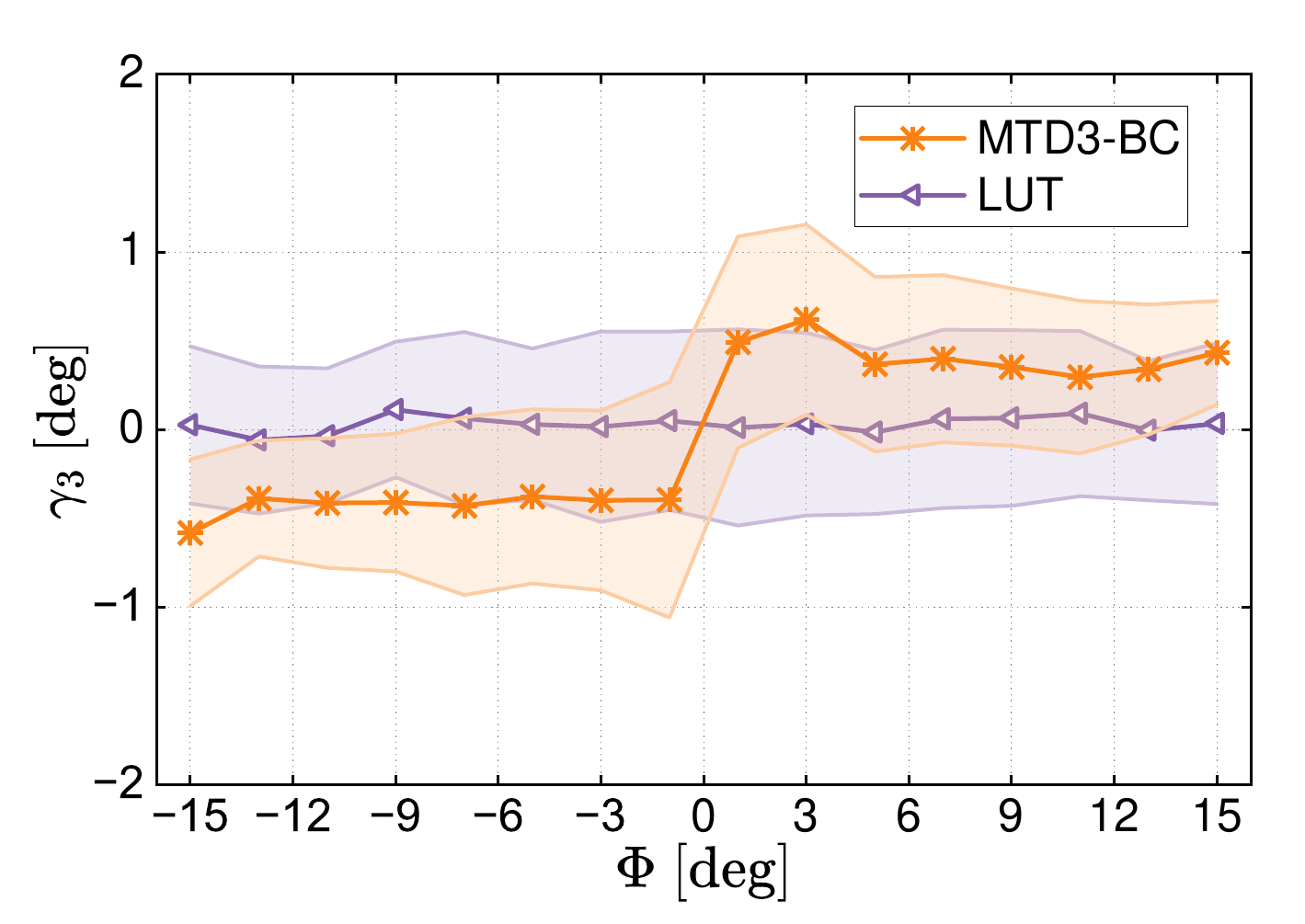}}
    \caption{Yaw angles of MTD3-BC and LUT under different wind directions.}
    \label{td3_lut_yaw}
\end{figure*}

Fig.~\ref{td3_lut_power} compares the power gains of MTD3-BC and LUT over the greedy strategy at the farm and individual turbine levels. At the farm level (Fig.~\ref{td3_lut_wf}), both MTD3-BC and LUT exhibit similar power gains, demonstrating their effectiveness in improving overall wind farm efficiency. However, at the turbine level, distinct differences emerge between the two methods. For WT1 (Fig.~\ref{td3_lut_wt1}), the LUT method exhibits a more pronounced power drop than MTD3-BC, particularly around $\Phi \in \left[-3^\circ,3^\circ\right]$, suggesting that the offline RL-based approach results in less power loss at the upstream turbine. Conversely, for WT2 (Fig.~\ref{td3_lut_wt2}), LUT yields a higher peak power gain compared to MTD3-BC, indicating that the LUT-based approach may be more aggressive in enhancing downstream turbine performance at certain wind angles. For WT3 (Fig.~\ref{td3_lut_wt3}), both methods show comparable trends. The smaller upstream power loss of MTD3-BC is likely due to the fact that the RL approach directly penalizes excessive control actions through the reward function (\ref{reward}), whereas the LUT formulation used here did not include such a criterion in its optimization, although this would clearly have been possible.  

The error band plot in Fig.~\ref{td3_lut_yaw} illustrates the yaw offset variations under the MTD3-BC and LUT control strategies for the three wind turbines. The mean yaw schedules of the two methods nearly coincide, which further confirms that the offline RL policy recovers, purely from data, a yaw strategy consistent with the model-based optimum. A further observation is that, near the sign-switching region $\Phi \in \left[-3^\circ,3^\circ\right]$, the upstream turbine WT1 exhibits a smaller dispersion of the yaw offset under MTD3-BC. As previously noted, this is again attributable to the penalization of excessive yaw actions in the reward function (\ref{reward}), reinforced by the action consistency term in (\ref{td3_policy}), which together promote smooth yaw transitions as the wind direction crosses the switching point. In addition, the offline RL policy applies a small but systematic yaw offset to the downstream turbine WT3; since its magnitude remains below $1^\circ$, its practical relevance should be interpreted with caution. Overall, these results indicate that the trade-off between the two competing objectives in (\ref{reward}) is achieved by the proposed MTD3-BC algorithm. 

\begin{figure}[!t]
    \centering
    \subfloat[Wind farm power generation under MTD3-BC, LUT and greedy control policies. \label{power}]{
        \includegraphics[width=3.1in]{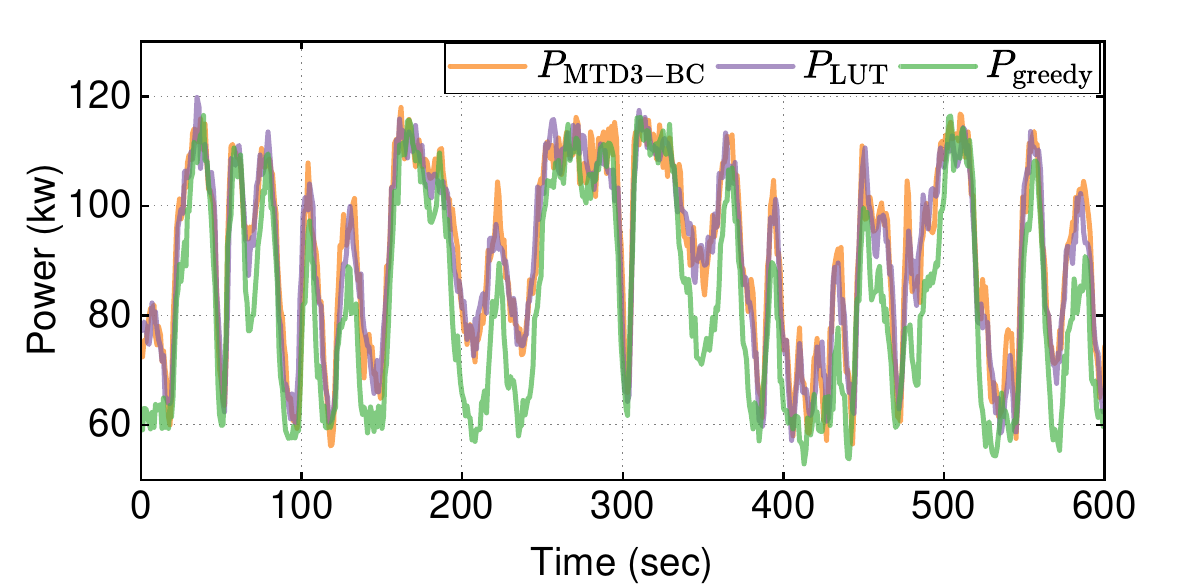}}  \\
    \subfloat[Time-varying turntable changes \label{wind}]{ 
         \includegraphics[width=3.1in]{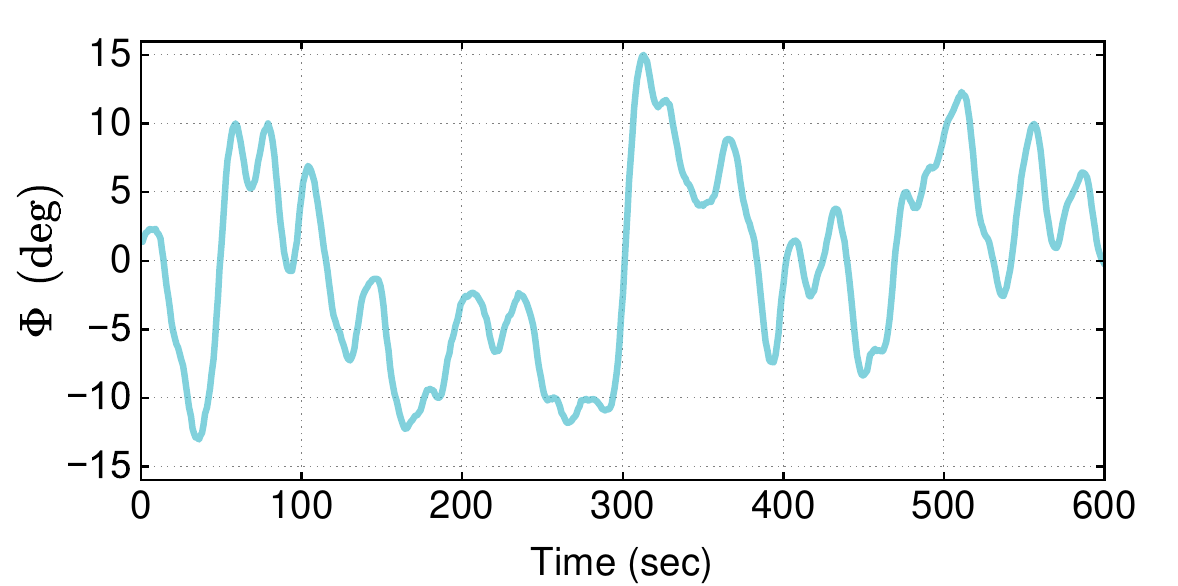}}
    \caption{Results of the fourth 10-minute wind profile.}
    \label{profile}
\end{figure}

In Fig.~\ref{profile}, we present the test results using the fourth 10-minute wind profile as an example.  Fig.~\ref{power} shows the time-varying farm-level generated power under the MTD3-BC, LUT, and Greedy control policies, while Fig.~\ref{wind} illustrates the time history of the turntable rotation angle. Fig.~\ref{profile} clearly demonstrates the effectiveness of our proposed offline RL algorithm in enhancing wind farm power production under time-varying wind directions.

\begin{remark}
The LUT-based and the proposed offline RL policies, although based on completely different approaches, achieve similar performance in this wind tunnel experiment; the residual differences are so small that they would probably not be measurable in a full-scale field deployment. The LUT approach relies on a model of the flow within the farm, whereas the RL policy does not. These results therefore show that, when the underlying wake model is of good quality, as is the case for the data-calibrated model used here under the controlled and repeatable conditions of the wind tunnel, the classical and widely studied LUT approach is difficult to beat and serves as a near-optimal benchmark. Viewed from the opposite direction, the same result indicates that the offline RL policy recovers this near-optimal behavior without access to any flow model. 
\end{remark}
\begin{remark}
Given this similarity in raw power capture, the practical advantages of the proposed framework lie elsewhere. A LUT provides yaw set-points only on a discrete grid of operating conditions, so interpolation is needed in between. Near wind directions where the optimal yaw offset changes sign, neighboring grid points can have yaw set-points with opposite signs. Interpolating between these points can therefore produce a yaw command close to zero, even though the optimal yaw offsets on either side may still have relatively large magnitudes. This can weaken the wake-steering effect and reduce the associated power gain. In addition, the LUT becomes increasingly harder to build as further variables, such as turbulence intensity, are added. The offline RL policy instead maps the measured conditions directly to continuous yaw commands. More importantly, the configuration of an operational wind farm is not static: turbines are curtailed, taken offline for maintenance, or suffer faults, and each such event alters the wake topology of the farm. A LUT must in principle be re-derived from a model updated to each new configuration, and pre-computing tables for all realistic availability combinations of a large farm quickly becomes impractical. The offline RL framework, in contrast, can potentially be fine-tuned directly on operational data logged under the new configuration, reusing the same training pipeline without model recalibration. Demonstrating this adaptability experimentally is beyond the scope of the present study, but the algorithmic foundation established here constitutes a first step in that direction.
\end{remark}
\begin{remark}
Although the MTD3-BC control policy achieves only marginally better performance than the PPO control algorithm, its computational and time costs are significantly lower. Compared to the online PPO method, our offline RL policy can infer good behavior from the provided dataset without exploring the environment, reducing the training and fine-tuning time for the MTD3-BC control policy to just approximately $5\%$ of that required for training the PPO control policy. This reduction in computational cost further enhances the scalability and real-world applicability of the proposed RL-based wind farm control strategy.
\end{remark}

\begin{remark}
The benefits of the presented offline RL method come at the price of training data: just as a model-based controller can only be as good as its model, an offline RL policy is limited by the coverage of its dataset. The present experiment demonstrates, at small scale, the workflow outlined in Section~\ref{se1}: the policy was bootstrapped from data logged under greedy and LUT controllers, then took over the yaw control of the farm, and subsequently was refined twice further using  data collected during its own operation.
\end{remark}
\section{Conclusion}
\label{se5}
This article introduces the MTD3-BC algorithm, an offline reinforcement learning solution for wind farm power maximization tasks that accounts for dynamic changes in wind flow direction. The algorithm is based on the standard TD3 actor-critic structure and enhances the policy learning objective by incorporating a behavior cloning term and an action consistency term. As a fully data-driven, model-free method, the proposed MTD3-BC algorithm can exploit existing wind farm datasets, enabling a more computationally efficient and time-saving training process compared to other online RL algorithms. Results from wind tunnel tests validate the effectiveness of the proposed algorithm in dynamic wind environments, achieving performance on par with a data-calibrated model-based wake-steering benchmark and a marginal improvement over the best behavior policy contained in the training dataset, at approximately 5\% of the training cost of an online PPO controller. This work marks the first successful experimental validation of an offline RL-based control strategy for wind farms, offering a promising pathway for real-world implementation to improve wind farm operational efficiency. In future work, we plan to further verify the effectiveness of the proposed offline RL method in commercial wind farm deployments.

\bibliographystyle{IEEEtran}
\bibliography{ref}

@article{gonzalez2012wake,
  title={Wake effect in wind farm performance: Steady-state and dynamic behavior},
  author={Gonz{\'a}lez-Longatt, Francisco and Wall, Peter and Terzija, Vladimir},
  journal={Renewable Energy},
  volume={39},
  number={1},
  pages={329--338},
  year={2012},
  publisher={Elsevier}
}

@article{shakoor2016wake,
  title={Wake effect modeling: A review of wind farm layout optimization using Jensen' s model},
  author={Shakoor, Rabia and Hassan, Mohammad Yusri and Raheem, Abdur and Wu, Yuan-Kang},
  journal={Renewable and Sustainable Energy Reviews},
  volume={58},
  pages={1048--1059},
  year={2016},
  publisher={Elsevier}
}

@article{howland2022collective,
  title={Collective wind farm operation based on a predictive model increases utility-scale energy production},
  author={Howland, Michael F and Quesada, Jes{\'u}s Bas and Mart{\'\i}nez, Juan Jos{\'e} Pena and Larra{\~n}aga, Felipe Palou and Yadav, Neeraj and Chawla, Jasvipul S and Sivaram, Varun and Dabiri, John O},
  journal={Nature Energy},
  volume={7},
  number={9},
  pages={818--827},
  year={2022},
  publisher={Nature Publishing Group UK London}
}

@article{dar2016windfarm,
  title={Wind farm power optimization using yaw angle control},
  author={Dar, Zamiyad and Kar, Koushik and Sahni, Onkar and Chow, Joe H},
  journal={IEEE Transactions on Sustainable Energy},
  volume={8},
  number={1},
  pages={104--116},
  year={2016},
  publisher={IEEE}
}

@article{howland2019wind,
  title={Wind farm power optimization through wake steering},
  author={Howland, Michael F and Lele, Sanjiva K and Dabiri, John O},
  journal={Proceedings of the National Academy of Sciences},
  volume={116},
  number={29},
  pages={14495--14500},
  year={2019},
  publisher={National Academy of Sciences}
}

@article{vali2017adjoint,
  title={Adjoint-based model predictive control of wind farms: Beyond the quasi steady-state power maximization},
  author={Vali, Mehdi and Petrovi{\'c}, Vlaho and Boersma, Sjoerd and van Wingerden, Jan-Willem and K{\"u}hn, Martin},
  journal={IFAC-PapersOnLine},
  volume={50},
  number={1},
  pages={4510--4515},
  year={2017},
  publisher={Elsevier}
}

@article{vali2019adjoint,
  title={Adjoint-based model predictive control for optimal energy extraction in waked wind farms},
  author={Vali, Mehdi and Petrovi{\'c}, Vlaho and Boersma, Sjoerd and van Wingerden, Jan-Willem and Pao, Lucy Y and K{\"u}hn, Martin},
  journal={Control Engineering Practice},
  volume={84},
  pages={48--62},
  year={2019},
  publisher={Elsevier}
}

@inproceedings{kheirabadi2021real,
  title={Real-time relocation of floating offshore wind turbines for power maximization using distributed economic model predictive control},
  author={Kheirabadi, Ali C and Nagamune, Ryozo},
  booktitle={2021 American Control Conference (ACC)},
  pages={3077--3081},
  year={2021},
  organization={IEEE}
}

@article{gionfra2019wind,
  title={Wind farm distributed PSO-based control for constrained power generation maximization},
  author={Gionfra, Nicolo and Sandou, Guillaume and Siguerdidjane, Houria and Faille, Damien and Loevenbruck, Philippe},
  journal={Renewable energy},
  volume={133},
  pages={103--117},
  year={2019},
  publisher={Elsevier}
}

@article{wang2016novel,
  title={A novel control strategy approach to optimally design a wind farm layout},
  author={Wang, Longyan and Tan, Andy and Gu, Yuantong},
  journal={Renewable energy},
  volume={95},
  pages={10--21},
  year={2016},
  publisher={Elsevier}
}

@article{park2016bayesian,
  title={Bayesian ascent: A data-driven optimization scheme for real-time control with application to wind farm power maximization},
  author={Park, Jinkyoo and Law, Kincho H},
  journal={IEEE Transactions on Control Systems Technology},
  volume={24},
  number={5},
  pages={1655--1668},
  year={2016},
  publisher={IEEE}
}

@article{wallace2024reinforcement,
  title={Reinforcement learning control of hypersonic vehicles and performance evaluations},
  author={Wallace, Brent A and Si, Jennie},
  journal={Journal of Guidance, Control, and Dynamics},
  volume={47},
  number={12},
  pages={2587--2600},
  year={2024},
  publisher={American Institute of Aeronautics and Astronautics}
}

@article{chai2022design,
  title={Design and experimental validation of deep reinforcement learning-based fast trajectory planning and control for mobile robot in unknown environment},
  author={Chai, Runqi and Niu, Hanlin and Carrasco, Joaquin and Arvin, Farshad and Yin, Hujun and Lennox, Barry},
  journal={IEEE Transactions on Neural Networks and Learning Systems},
  volume={35},
  number={4},
  pages={5778--5792},
  year={2022},
  publisher={IEEE}
}

@article{meng2024online,
  title={An online reinforcement learning-based energy management strategy for microgrids with centralized control},
  author={Meng, Qinglin and Hussain, Sheharyar and Luo, Fengzhang and Wang, Zhongguan and Jin, Xiaolong},
  journal={IEEE Transactions on Industry Applications},
  year={2024},
  publisher={IEEE}
}

@article{zhao2020cooperative,
  title={Cooperative wind farm control with deep reinforcement learning and knowledge-assisted learning},
  author={Zhao, Huan and Zhao, Junhua and Qiu, Jing and Liang, Gaoqi and Dong, Zhao Yang},
  journal={IEEE Transactions on Industrial Informatics},
  volume={16},
  number={11},
  pages={6912--6921},
  year={2020},
  publisher={IEEE}
}

@article{dong2023reinforcement,
  title={Reinforcement learning-based wind farm control: Toward large farm applications via automatic grouping and transfer learning},
  author={Dong, Hongyang and Zhao, Xiaowei},
  journal={IEEE Transactions on Industrial Informatics},
  volume={19},
  number={12},
  pages={11833--11845},
  year={2023},
  publisher={IEEE}
}

@article{huang2024reinforcement,
  title={Reinforcement Learning-Based Multiobjective Control of Grid-Connected Wind Farms},
  author={Huang, Yubo and Zhao, Xiaowei},
  journal={IEEE Transactions on Industrial Informatics},
  year={2024},
  publisher={IEEE}
}

@article{schulman2017proximal,
  title={Proximal policy optimization algorithms},
  author={Schulman, John and Wolski, Filip and Dhariwal, Prafulla and Radford, Alec and Klimov, Oleg},
  journal={arXiv preprint arXiv:1707.06347},
  year={2017}
}

@article{campagnolo2020wind,
  title     = {Wind tunnel testing of wake steering with dynamic wind direction changes},
  author    = {Campagnolo, Filippo and Weber, Robin and Schreiber, Johannes and Bottasso, Carlo L.},
  journal   = {Wind Energy Science},
  volume    = {5},
  number    = {4},
  pages     = {1273--1295},
  year      = {2020},
  publisher = {Copernicus Publications},
  doi       = {10.5194/wes-5-1273-2020}
}

@article{fujimoto2021minimalist,
  title={A minimalist approach to offline reinforcement learning},
  author={Fujimoto, Scott and Gu, Shixiang Shane},
  journal={Advances in neural information processing systems},
  volume={34},
  pages={20132--20145},
  year={2021}
}

@article{bottasso2014wind,
  title={Wind tunnel testing of scaled wind turbine models: Beyond aerodynamics},
  author={Bottasso, Carlo L and Campagnolo, Filippo and Petrovi{\'c}, Vlaho},
  journal={Journal of wind engineering and industrial aerodynamics},
  volume={127},
  pages={11--28},
  year={2014},
  publisher={Elsevier}
}

@inproceedings{fujimoto2018addressing,
  title={Addressing function approximation error in actor-critic methods},
  author={Fujimoto, Scott and Hoof, Herke and Meger, David},
  booktitle={International conference on machine learning},
  pages={1587--1596},
  year={2018},
  organization={PMLR}
}

@article{lillicrap2015continuous,
  title={Continuous control with deep reinforcement learning},
  author={Lillicrap, Timothy P and Hunt, Jonathan J and Pritzel, Alexander and Heess, Nicolas and Erez, Tom and Tassa, Yuval and Silver, David and Wierstra, Daan},
  journal={arXiv preprint arXiv:1509.02971},
  year={2015}
}

@inproceedings{campagnolo2022further,
  title={Further calibration and validation of FLORIS with wind tunnel data},
  author={Campagnolo, Filippo and Im{\v{s}}irovi{\'c}, Lejla and Braunbehrens, Robert and Bottasso, Carlo L},
  booktitle={Journal of Physics: Conference Series},
  volume={2265},
  number={2},
  pages={022019},
  year={2022},
  organization={IOP Publishing}
}

@article{bottasso2022wind,
  title={Wind tunnel testing of wind turbines and farms},
  author={Bottasso, Carlo L and Campagnolo, Filippo},
  journal={Handbook of wind energy aerodynamics},
  pages={1077--1126},
  year={2022},
  publisher={Springer}
}


\begin{IEEEbiography}[{\includegraphics[width=1in,height=1.25in,clip,keepaspectratio]{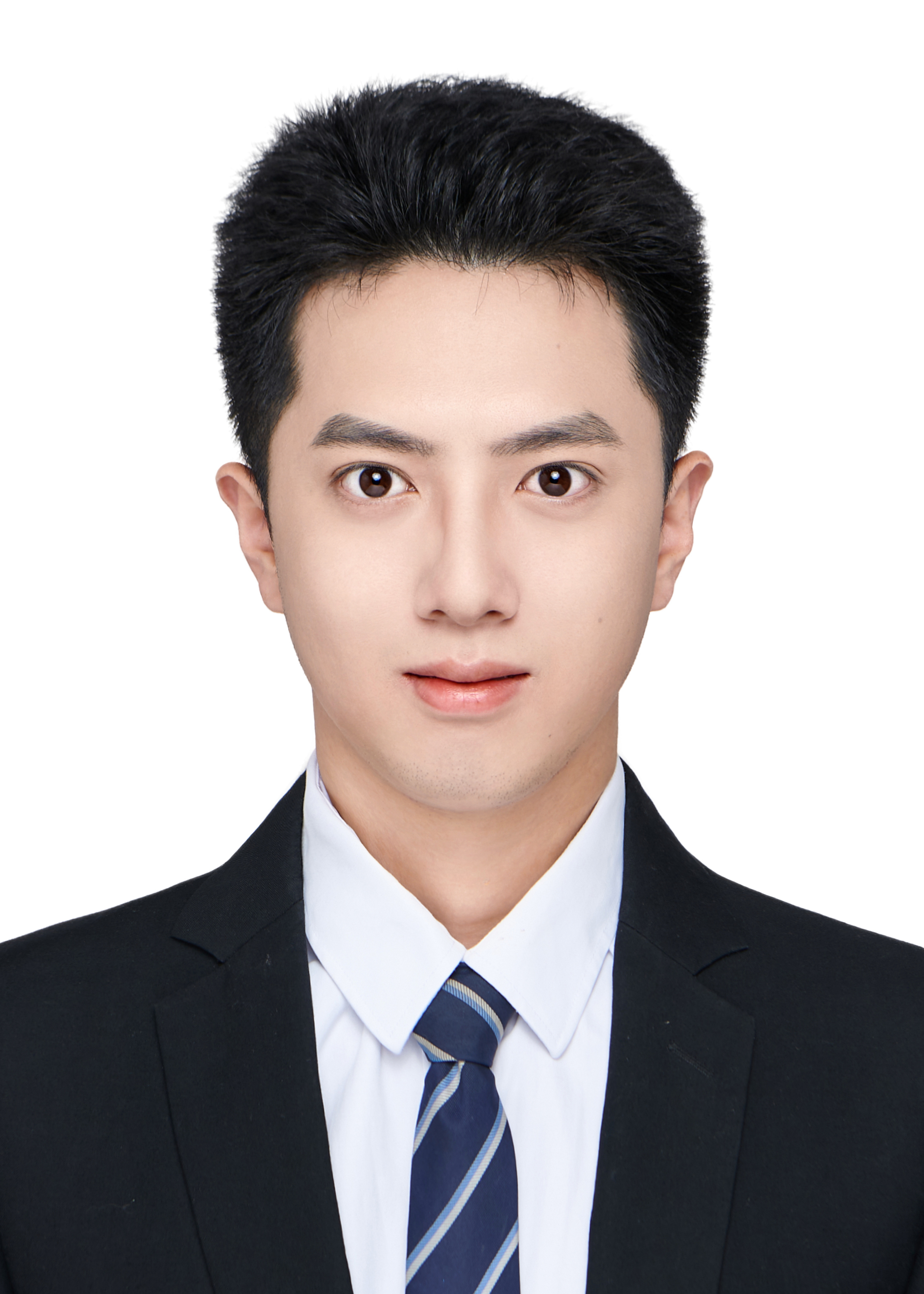}}]{Yuhan Su} was born in Anqing, China. He
received the M.Eng. degree in electronics and informatics from Xiamen University,
Xiamen, China, in 2024. He is currently working toward the Ph.D. degree in engineering with the University of Warwick, Coventry, U.K., where he is also a Marie Curie Early Stage Researcher. His research interests include control theory and reinforcement learning, with applications to offshore renewable energy systems.
\end{IEEEbiography}

\begin{IEEEbiography}[{\includegraphics[width=1in,height=1.25in,clip,keepaspectratio]{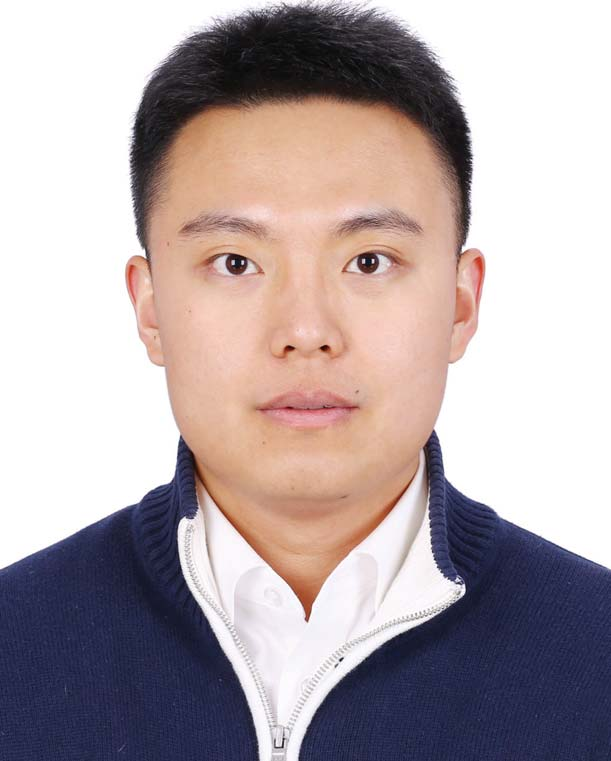}}]{Hongyang Dong}  is an Assistant Professor with the School of Engineering, University of Warwick, Coventry, U.K. He was a Research Fellow of
machine learning and intelligent control with the University of Warwick, from 2019 to 2022, before he became an Assistant Professor, in November 2022. His current research interests include control theories and machine learning methods with their applications in complex systems, including offshore renewable energy systems and autonomous systems.
\end{IEEEbiography}

\begin{IEEEbiography}[{\includegraphics[width=1in,height=1.25in,clip,keepaspectratio]{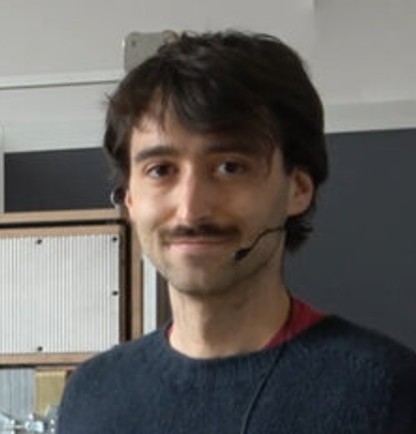}}]{Simone Tamaro} was born in Trieste, Italy. He received the B.Sc. degree in mechanical engineering from the University of Trieste, Italy, in 2017; the M.Sc. degrees in aerospace engineering and wind engineering from Delft University of Technology, the Netherlands, and the Technical University of Denmark, Denmark, in 2019; and a Research Master’s degree from the von Karman Institute for Fluid Dynamics, Belgium, in 2020. He is currently a Ph.D. Candidate at the Wind Energy Institute at the Technical University of Munich, Garching b. Munich, Germany. His research focuses on computational fluid dynamics, wind farm control, experimental aerodynamics and turbulence.
\end{IEEEbiography}

\begin{IEEEbiography}[{\includegraphics[width=1in,height=1.25in,clip,keepaspectratio]{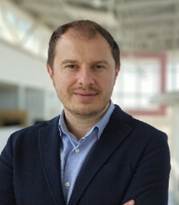}}]{Filippo Campagnolo} was born in Vicenza, Italy. He received the B.Sc. degree in aerospace engineering from the University of Padova in 2004, and the M.Sc. and Ph.D. degrees in aeronautical engineering and rotary-wing aircraft, respectively, from Politecnico di Milano in 2008 and 2013. He is currently a Senior Researcher and Lecturer with the Wind Energy Institute, Technical University of Munich, Garching b. Munich, Germany. He has authored or co-authored over 120 publications on the experimental and computational design and control of wind turbines and wind farms.

\end{IEEEbiography}

\begin{IEEEbiography}[{\includegraphics[width=1in,height=1.25in,clip,keepaspectratio]{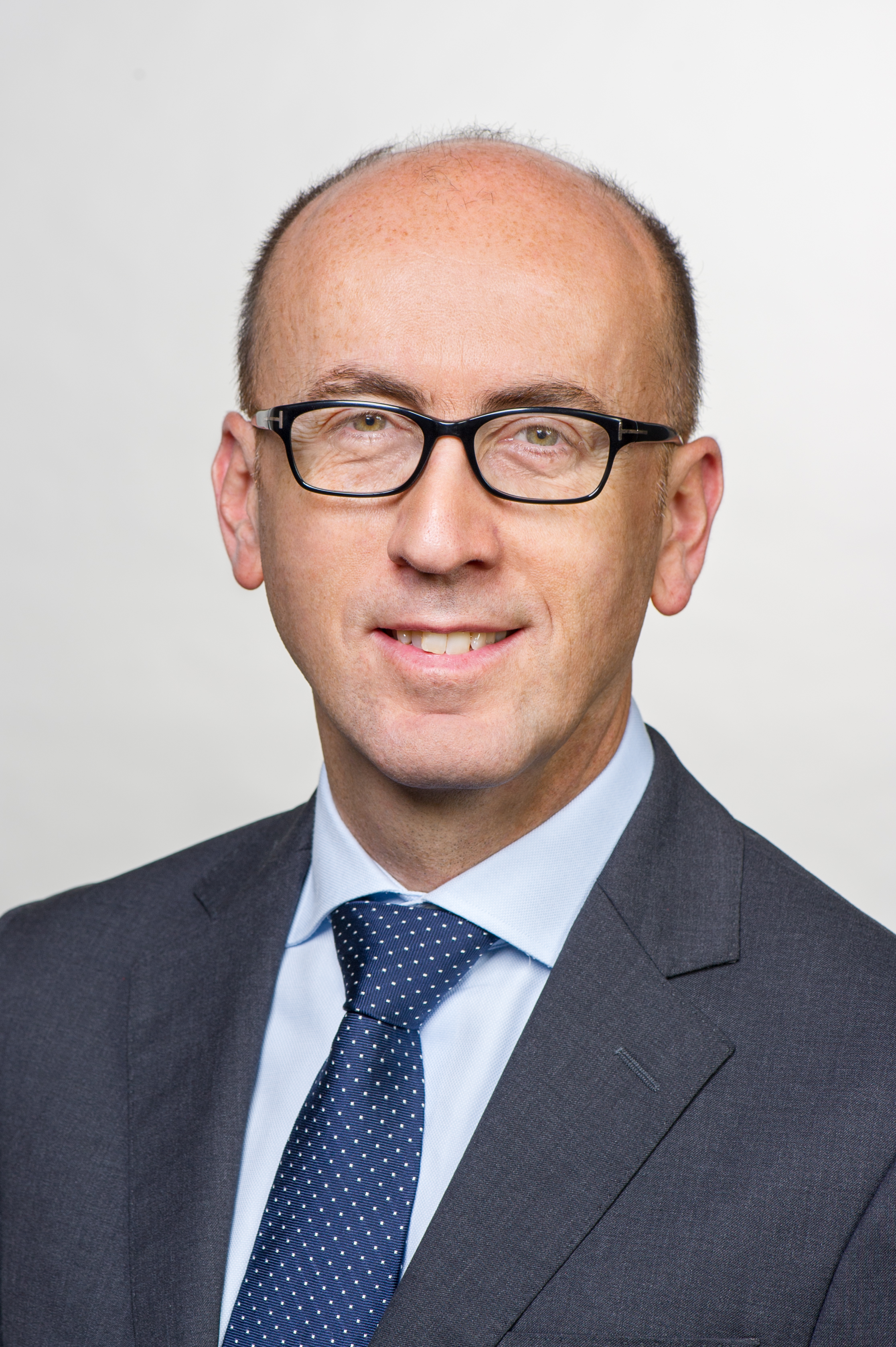}}]{Carlo L. Bottasso}
was born in Milano, Italy. He received a M.Sc. and Ph.D. in aerospace engineering from Politecnico di Milano, Italy, in 1987 and 1993 respectively. He holds the Chair of Wind Energy at the Technical University of Munich (TUM), where he is also the founding director of the Wind Energy Institute. Prof. Bottasso has been the President and Vice-President of the European Academy of Wind Energy (EAWE), and he is the Editor in Chief of the Wind Energy Science journal. In 2023, he was honored with the EAWE Scientific Award.
\end{IEEEbiography}

\begin{IEEEbiography}[{\includegraphics[width=1in,height=1.25in,clip,keepaspectratio]{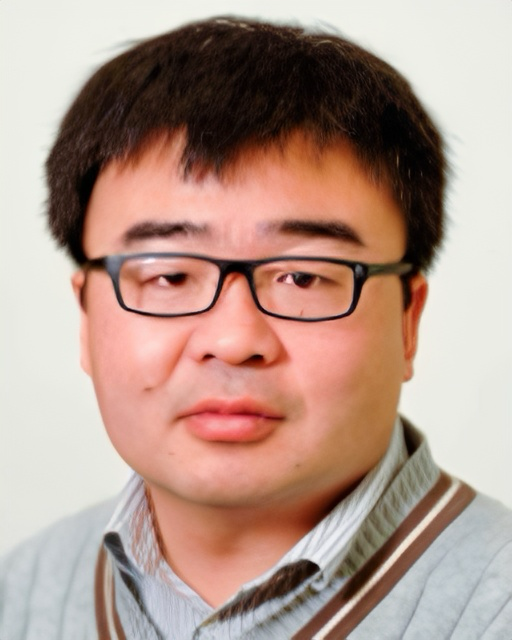}}]{Xiaowei Zhao} (Member, IEEE) received the Ph.D.
degree in control theory from Imperial College
London, London, U.K., in 2010.
He is a Professor of control engineering with
the School of Engineering, University of Warwick,
Coventry, U.K. He was a Post-Doctoral Researcher
with the University of Oxford, Oxford, U.K., for
three years before joining the University of Warwick,
in 2013. His main research interests include control
theory and machine learning with applications in
offshore renewable energy systems, smart grids, and
autonomous systems.
\end{IEEEbiography}

\end{document}